\pdfoutput=1
\documentclass[11pt]{article}
\usepackage[final]{acl}
\usepackage{times}
\usepackage{latexsym}
\usepackage[T1]{fontenc}
\usepackage[utf8]{inputenc}
\usepackage{microtype}
\usepackage{inconsolata}
\usepackage{graphicx}
\usepackage{caption}
\usepackage{subcaption}
\usepackage{amsmath,amsfonts,amssymb}
\usepackage{textcomp}
\usepackage{url}
\usepackage{booktabs}
\usepackage{multirow}
\usepackage{arydshln}
\usepackage{makecell}
\usepackage{multirow}
\usepackage{pifont}
\usepackage[linesnumbered,ruled,vlined]{algorithm2e}
\usepackage{color}
\usepackage{xcolor,pifont}

\newcommand*\colourcross[1]{%
  \expandafter\newcommand\csname #1cross\endcsname{\textcolor{#1!75}{\ding{56}}}%
}
\colourcross{red}
\newcommand*\colourcheck[1]{%
  \expandafter\newcommand\csname #1check\endcsname{\textcolor{#1!75}{\ding{52}}}%
}
\colourcheck{green}
\newcommand\best{\def\argii{}\docommandbest}
\def\docommandbest#1 {\colorbox{cyan!20}{#1} \let\next\argii}
\def\argii{\let\next\docommandbest}

\newcommand\better{\def\argii{}\docommandbetter}
\def\docommandbetter#1 {\colorbox{green!20}{#1} \let\next\argii}
\def\argii{\let\next\docommandbetter}

\newcommand{\compactbest}[1]{\setlength{\fboxsep}{1pt}\colorbox{cyan!20}{#1}}
\newcommand{\compactbetter}[1]{\setlength{\fboxsep}{1pt}\colorbox{green! 20}{#1}}

\newcommand{\ve}[1]{\boldsymbol{#1}}

\newcommand{\set}[1]{\mathbb{#1}}

\title{Registering Source Tokens to Target Language Spaces \\ in Multilingual Neural Machine Translation}

\author{
\textbf{Zhi Qu\thanks{This work was done during the first author’s internship at Advanced Speech Translation Research and Development Promotion Center, National Institute of Information and Communications Technology, Kyoto, Japan.}$^1$ }
\textbf{Yiran Wang$^2$ }
\textbf{Jiannan Mao$^2$}
\\
\textbf{Chenchen Ding$^1$$^2$}
\textbf{Hideki Tanaka$^2$ }
\textbf{Masao Utiyama$^2$ }
\textbf{Taro Watanabe$^1$}
  \\
  $^1$Nara Institute of Science and Technology, Japan.
  \\
  $^2$National Institute of Information and Communications Technology, Japan.
  \\
  \texttt{qu.zhi.pv5@is.naist.jp}
}

\begin{document}
\maketitle
\begin{abstract}
The multilingual neural machine translation (MNMT) aims for arbitrary translations across multiple languages.
Although MNMT-specific models trained on parallel data offer low costs in training and deployment, their performance consistently lags behind that of large language models (LLMs).
In this work, we introduce \textbf{registering}, a novel method that enables a small MNMT-specific model to compete with LLMs.
Specifically, we insert a set of artificial tokens specifying the target language, called registers, into the input sequence between the source and target tokens.
By modifying the attention mask, the target token generation only pays attention to the activation of registers, representing the source tokens in the target language space.
Experiments on EC-40, a large-scale benchmark, show that our method advances the state-of-the-art of MNMT.
We further pre-train two models, namely MITRE (\textbf{m}ult\textbf{i}lingual \textbf{t}ranslation with \textbf{re}gisters), by 9.3 billion sentence pairs across 24 languages collected from public corpora.
One of them, MITRE-913M, outperforms NLLB-3.3B, achieves comparable performance with commercial LLMs, and shows strong adaptability in fine-tuning.
Finally, we open-source our models to facilitate further research and development in MNMT: \url{https://github.com/zhiqu22/mitre}.
\end{abstract}

\section{Introduction}
Multilingual neural machine translation (MNMT) aims to enable arbitrary translations across multiple languages.
Traditionally, training models specific to MNMT using parallel data was highly appealing, not only because such MNMT-specific models maintain a minimal number of parameters \cite{ZeroMNMT2016, m2m, nllb}, but also due to their potential for zero-shot translation, i.e., translating language pairs unseen during training, which helps address data scarcity in certain translation directions \cite{GooglesMNMT, m2m, massive-2020}.
However, the current mainstream solution for MNMT relies on large language models (LLMs), as the performance of MNMT-specific models has lagged behind that of LLMs \cite{mnmtVSllm, ftllm}.
Recent analyses \cite{target-off, zero-2023} identify the off-target problem, i.e., translations fail to reach the intended target language, as a key factor causing the under-performance of MNMT-specific models.
Moreover, \citet{subword-2020,target-off} show that constraining the target token generation to the target language space can alleviate the off-target problem.

\begin{figure}[t]
    \centering
        \includegraphics[width=0.8\linewidth]{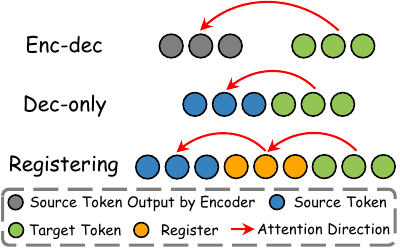}
    \caption{Illustration of the attention view among different architectures.
    "Token" refers to the representation corresponding to the token.}
    \label{fig:attention_direction}
\end{figure}

In this work, we present \textbf{registering}, a simple yet effective method designed for MNMT-specific models based on the decoder-only architecture (Dec-only) without introducing additional parameters.
Specifically, we insert a set of artificial tokens between the source and target tokens, called registers, which indicate only the target language without any semantics.
The registers are designed to have the same length as the source tokens, because each register is expected to capture the semantics of its positionally-aligned source token and then represent it in the target language space.
As illustrated in Figure \ref{fig:attention_direction}, by modifying the attention mask, the generation of target tokens no longer follows the attention mechanism of encoder-decoder (Enc-dec) or Dec-only architectures. Instead, it relies solely on the registers located in the representational space of the intended target language.

We conduct two sets of experiments, evaluated with four automatic metrics: spBLEU \cite{nllb}, chrF++ \cite{chrf, chrf++}, COMET \cite{comet}, and off-target ratio \cite{massive-2020}.
First, we experiment with EC-40 \cite{zero-2023}, a large-scale benchmark designed to assess zero-shot translation capability.
Experimental results show that, compared to strong baselines, our method improves spBLEU scores by up to 71\% on average across 1,640 directions with fewer parameters and drastically reduces the off-target ratio from 26.69\% to 3.65\%.
Second, we collect 9.3 billion sentence pairs across 24 languages by sampling from the NLLB open dataset \cite{nllb} with the bridge language strategy \cite{m2m}.
We then pre-train two models, MITRE-466M and MITRE-913M (\textbf{m}ult\textbf{i}lingual \textbf{t}ranslation with \textbf{re}gisters).
One of them, MITRE-913M, not only outperforms NLLB-3.3B \cite{nllb} and GPT-3.5 Turbo \cite{gpt3} but also achieves competitive performance with GPT-4o mini \cite{gpt4}.
Also, we fine-tune the pre-trained models with full parameters and LoRA \cite{lora-2022} in three distinct scenarios, demonstrating the superior adaptability of MITRE in fine-tuning.
Finally, by analyzing the attention mechanisms and the representation distribution in translation instances at the token level, we confirm that the register mirrors the corresponding source token in the target language space.

\section{Related Work}\label{section:related}
\citet{GooglesMNMT} proposed adding a language tag as a translation instruction at the beginning of the input sequence, marking the beginning of MNMT-specific models.
Recent analyses \cite{target-off, zero-2023} show that addressing the off-target problem is key to improving zero-shot translation.
Early works \cite{subword-2020, adaptingZero} tried to use language-specific dictionaries or components to isolate generation across languages, but this was costly and hindered knowledge sharing \cite{share-2021}.
As a compromise, \citet{target-off} proposed adding language-specific subsets to a shared dictionary to mitigate the off-target problem.
Beyond the explicit addition of language-specific parameters, optimizing internal representations implicitly improves zero-shot translation.
Specifically, aligning semantic information across languages \cite{Constras-2021, aganostic-2024} and strengthening the translation instruction toward the target language \cite{transfer-2023, transfer, lcs-2024} help mitigate the off-target problem.
Our proposed method is a combination of explicit and implicit strategies, as it separates generation-related representations by language.

In methodology, the translation instruction used in MNMT-specific models \cite{GooglesMNMT} is similar to the artificial tokens in prefix-tuning \cite{prefix-2021}.
In fact, we are methodologically inspired by gisting, a variation of prefix-tuning proposed by \citet{gist}.
Specifically, they modified the attention mask to compress information into a set of artificial tokens, used in generation to eliminate the need for the original sequence.
However, the difference in concept is that our proposed method, registering, aims to transfer each source token's semantics into the positionally-aligned register, which can be regarded as a representation-level container pointing to the target language.
In other words, registering is conceptually similar to chain-of-thought \cite{cot}, where the process represents "rethinking" the source tokens from the perspective of the target language.

Additionally, given that LLMs already exhibit strong MNMT capabilities \cite{mnmtVSllm}, fine-tuning LLMs into MNMT-specific models has become a popular direction of exploration.
However, the results in this direction are still limited, such as performance still falling behind commercial LLMs \cite{bigtranslate} and fewer supported languages, e.g., 5 in \citet{ftllm} and 10 in \citet{tower_llm_2024}.
In this work, we demonstrate the potential of directly training an MNMT-specific model with parallel data only, aiming to drive further discussion on MNMT.

\begin{figure*}[t]
    \centering
    \includegraphics[width=\linewidth]{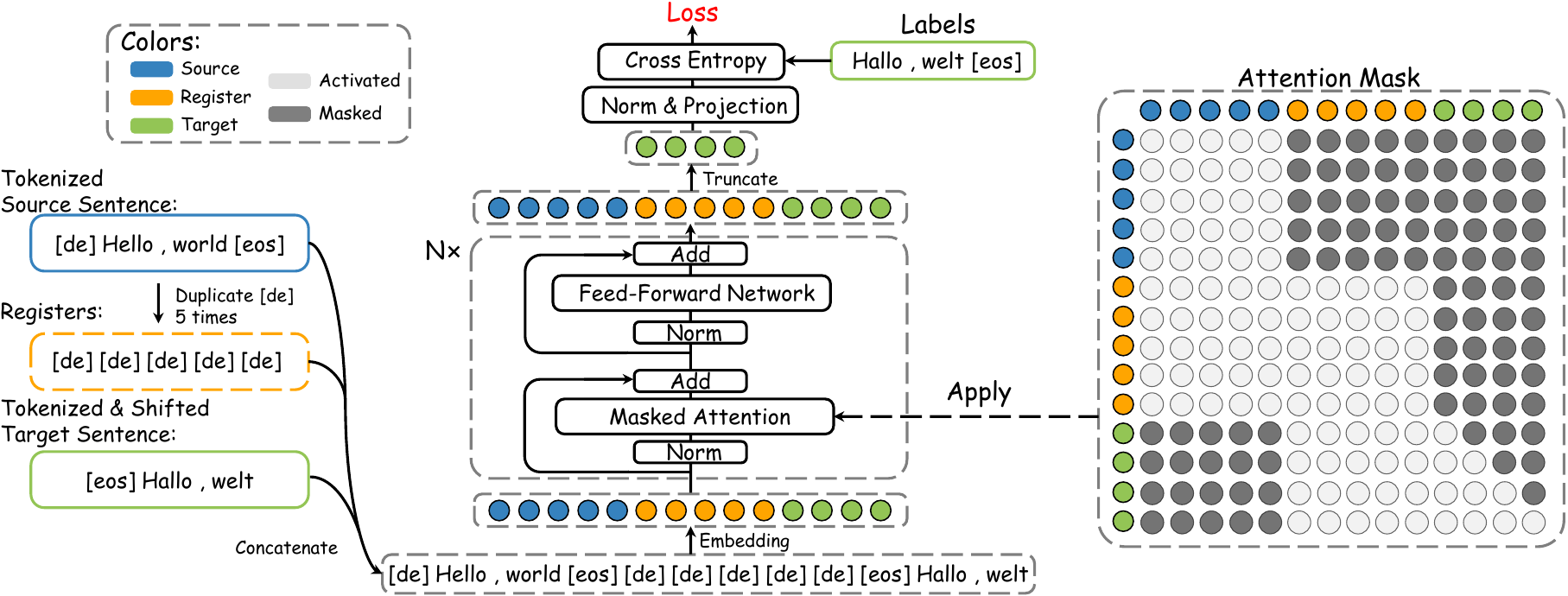}
    \caption{Illustration of registering.
    The example depicts a translation from English to German.
    The illustrated model stacks N layers, each following the Transformer decoder layer structure \cite{transformer} with pre-normalization \cite{pre_norm}.
    Notably, each circle represents a token and its representation in the generation.
    }
    \label{fig:main}
\end{figure*}

\section{Multilingual Translation With Registers}
\subsection{Multilingual Neural Machine Translation}
Given a multilingual corpus $\set{C}$ spanning multiple translation directions, each instance in $\set{C}$ is defined as $(\ve{x}, \ve{y})$, where $\ve{x}$ consists of a set of source tokens $\ve{x} = x_1, \ldots, x_I$ and a set of target tokens $\ve{y} = y_1, \ldots, y_J$. 
Also, we introduce a set of language tags $\mathbb{L}=\{l_1, \ldots, l_K\}$, which are artificial tokens, each corresponding to one of the $K$ languages in $\set{C}$.
Following \citet{GooglesMNMT, TagMatter_2021}, we add a tag indicating the language of $\ve{y}$ at the beginning of $\ve{x}$ as the translation instruction for multilingual neural machine translation (MNMT), denoted by $l_{\ve{y}}$.
Consequently, the input fed into the MNMT-specific model becomes $\ve{x}' = l_{\ve{y}}, x_1, \ldots, x_I$.
Formally, we train the model over all instances of $\set{C}$ by optimizing the following cross entropy loss:
\begin{equation}\label{eq:ce}
\mathcal{L}_{\text{ce}}= -\sum_{
  \ve{x}', \ve{y} \in \set{C}}{\sum_{j=1}^{J}
    {
      \log {p(y_{j} \mid \ve{x}', \ve{y}_{<j})
    },
  }
}
\end{equation}
where $p(y_{j} \mid \ve{x}', \ve{y}_{<j})$ represents the probability for generating $y_{j}$ by MNMT-specific model.
The current state-of-the-art models \cite{m2m, nllb} utilize the encoder-decoder architecture (Enc-dec), where the generation of $y_{j}$ can be expressed as:
\begin{equation}\label{eq:enc_dec}
y_{j} = \operatorname{decoder}(\operatorname{encoder}(\ve{x}'), \ve{y}_{<j}).
\end{equation}
Also, \citet{decoderonlymt, decoderonly_2022} show that the MNMT-specific model can be implemented with a decoder-only architecture (Dec-only).
In this setup\footnote{We follow \citet{decoderonlymt} to train a Dec-only MNMT-specific model with Equation \ref{eq:ce} rather than using a language modeling loss \cite{gpt1}.}, the generation can be described as:
\begin{equation}\label{eq:dec_only}
y_{j} = \operatorname{decoder}(\ve{x}', \ve{y}_{<j}).
\end{equation}

\subsection{Registering}\label{section:registering}
\citet{spurious_2019, tdo} suggest that $l_{\ve{y}}$ is a key factor causing the off-target problem.
Specifically, although $l_{\ve{y}}$ is a translation instruction distinctly representing the target language, generation cannot strictly depend on $l_{\ve{y}}$ due to the dilution of other source tokens in the attention mechanism.
Therefore, registering is proposed to address this problem by constraining the generation within the target language space, where all registers used in generation have the same function as $l_{\ve{y}}$.

Given a Dec-only model\footnote{Intuitively, Dec-only is more parameter-efficient than Enc-dec, as separate components in the latter process source and target tokens. Therefore, given that registering can address the off-target problem, Dec-only is the preferable choice. Supporting experiments are provided in Appendix \ref{appendix:compare}.}, we begin by initializing a set of artificial tokens corresponding to the target language, denoted by $\ve{r} = r_1, \ldots, r_{I+1}$, matching the length of $\ve{x}'$.
Notably, since $l_{\ve{y}}$ has the same function as we design for $r$, $\ve{r}$ is initialized by duplicating $l_{\ve{y}}$.
We then insert $\ve{r}$ into the input sequence between $\ve{x}'$ and $\ve{y}$, thus reformulating the generation process of Equation \ref{eq:dec_only} as:
\begin{equation}\label{eq:register}
y_{j} = \operatorname{decoder}(\ve{x}', \ve{r}, \ve{y}_{<j}).
\end{equation}

\begin{table*}[!ht]
    \centering
    \resizebox{\textwidth}{!}{
    \begin{tabular}{cclccccccccccccc}
    \toprule
        ~  & ~ & ~ &  \multicolumn{2}{c}{High} & \multicolumn{2}{c}{Med} & \multicolumn{2}{c}{Low} & \multicolumn{2}{c}{Extra Low} & ~ & ~ & ~ & ~ \\ 
    \midrule
        ~ & \#params & Method & $\rightarrow$ & $\leftarrow$ & $\rightarrow$ & $\leftarrow$ & $\rightarrow$ & $\leftarrow$ & $\rightarrow$ & $\leftarrow$ & sup. & zero. & avg. & off.(\%) \\ 
    \midrule
        \multirow{4}{*}{Enc-dec} & \multirow{3}{*}{242M} & vanilla & 9.46  & 11.05  & 7.49  & 9.80  & 5.03  & 3.95  & 5.41  & 2.59  & 29.06  & 5.86  & 6.99 & 48.40 \\ 
        ~ & ~ & \hspace{0.4em}+CL & 14.21  & 14.19  & 12.19  & 14.18  & 7.89  & 7.55  & 7.66  & 6.04  & 29.03  & 9.74  & 10.68 & 19.08 \\ 
        ~ & ~ & \hspace{0.4em}+LCS & 10.44  & 13.67  & 9.34  & 13.17  & 8.73  & 6.07  & 8.49  & 4.10  & \textbf{29.18}  & 8.35  & 9.37 & 22.43 \\
        ~ & 259M & \hspace{0.4em}+LAVS & 9.71 & 12.22 & 7.74 & 11.19 & 5.71 & 3.98 & 6.37 & 2.99 & 29.07 & 6.40 & 7.54 & 42.98 \\
    \cdashline{1-15}
        \multirow{3}{*}{Dec-only} & \multirow{3}{*}{217M} & vanilla & 11.57  & 11.51  & 9.48  & 11.03  & 6.01  & 6.06  & 5.74  & 4.20  & 28.61  & 7.30  & 8.34 & 22.21 \\ 
        ~ & ~ & \hspace{0.4em}+TDO & 13.33  & 13.50  & 10.53  & 12.75  & 7.13  & 6.88  & 6.74  & 4.60  & 28.84  & 8.62  & 9.61 & 27.18 \\ 
        ~ & ~ & \hspace{0.4em}+Ours & \textbf{15.43}  & \textbf{15.46}  & \textbf{13.88}  & \textbf{14.62}  & \textbf{8.94}  & \textbf{8.99}  & \textbf{8.76}  & \textbf{7.94}  & 28.90  & \textbf{11.05}  & \textbf{11.92} & \textbf{4.65} \\
    \midrule
        \multirow{4}{*}{Enc-dec} & \multirow{3}{*}{418M} & vanilla & 12.66  & 15.02  & 10.86  & 14.50  & 7.40  & 5.04  & 7.12  & 3.47  & 30.28  & 8.64  & 9.69 & 26.69 \\ 
        ~ & ~ & \hspace{0.4em}+CL & 15.89  & 15.97  & 13.67  & 16.15  & 8.36  & 8.16  & 8.32  & 5.96  & \textbf{30.54}  & 10.79  & 11.76 & 19.99 \\ 
        ~ & ~ & \hspace{0.4em}+LCS & 10.79  & 16.19  & 10.00  & 15.31  & 9.99  & 5.39  & 9.41  & 3.70  & 30.33  & 9.25  & 10.28 & 23.47 \\ 
        ~ & 430M & \hspace{0.4em}+LAVS & 14.03 & 16.39 & 12.50 & 16.31 & 8.61 & 6.31 & 8.45 & 3.57 & 30.20 & 10.03 & 11.01 & 21.73 \\ 
    \cdashline{1-15}
        \multirow{3}{*}{Dec-only} & \multirow{3}{*}{368M} & vanilla & 14.37  & 15.07  & 12.25  & 15.02  & 8.27  & 7.40  & 7.71  & 5.11  & 29.97  & 9.84  & 10.82 & 19.01 \\ 
        ~ & ~ & \hspace{0.4em}+TDO & 15.27  & 15.79  & 12.83  & 15.56  & 8.44  & 7.96  & 8.15  & 5.40  & 30.23  & 10.40  & 11.37 & 23.14 \\ 
        ~ & ~ & \hspace{0.4em}+Ours & \textbf{16.81}  & \textbf{16.98}  & \textbf{15.25}  & \textbf{16.57}  & \textbf{10.10}  & \textbf{9.88}  & \textbf{9.64}  & \textbf{8.37}  & 29.88  & \textbf{12.26}  & \textbf{13.12} & \textbf{3.65} \\ 
    \bottomrule
    \end{tabular}
    }
    \vspace{-0.3em}
    \caption{Averaged spBLEU scores of results on EC-40, the last column (off.) lists the off-target ratio averaged from all directions, the scores of chrF++ and COMET are reported in Tables \ref{tab:result_ec40_chrf} and \ref{tab:result_ec40_comet}, as discussed in Appendix \ref{appendix:results_ec40}.
    We report scores by grouping languages that have the same resource tier.
    Then, $\rightarrow$ includes directions translating from the corresponding group to languages out of this group, and $\leftarrow$ includes directions translating to the corresponding group.
    sup., zero, and avg. abbreviate the average of supervised translations, the average of zero-shot translations, and the average of all translations, respectively.
    The best score in each column of a block is in bold.
    }
    \vspace{-0.7em}
    \label{tab:result_ec40}
\end{table*}
The key step of registering is modifying the Transformer attention mask \cite{transformer} to remove source tokens from the view of target tokens.
As shown in Figure \ref{fig:main}, we initialize the attention mask based on the prefix Dec-only scheme \cite{unilm}, where the source tokens compute attention for each other bidirectionally, while the target tokens only compute attention for the previous ones.
We then adjust the mask to control token-level representation according to the following rules:
(1) $\ve{r}$ pays attention to $\ve{x}'$;
(2) $\ve{r}$ computes attention bidirectionally within $\ve{r}$;
(3) $y_{j}$ pays attention to $\ve{r}$ and $\ve{y}_{<j}$.
With this design, the generation of $\ve{y}$ can solely rely on the activation of $\ve{r}$, where the activation of $r_i$ not only functions as a representational container of the target language but also carries the semantics of $x_i$.
As a result, the generation is strictly constrained to the target language space to effectively address the off-target problem.

\section{Experiment: On Benchmark}\label{section:benchmark}

\subsection{Dataset and Evaluation}\label{section:benchmark_dataset}
We conduct the first set of experiments on a large zero-shot translation benchmark, EC-40 \cite{zero-2023}, consisting of 120 million translation instances spanning 41 languages across five language families in the training data\footnote{Details of EC-40 are described in Appendix \ref{appendix:ec40}.}.
Except for English, each family includes eight languages, categorized into four resource tiers, namely, High, Medium, Low, and Extra Low, corresponding to 5M, 1M, 100K, and 50K sentence pairs, respectively.
For testing the zero-shot translation capability, all training directions in EC-40 involve English, either as the source or target language, resulting in 80 supervised directions.
Then, we evaluate the trained models on all supervised and zero-shot directions, i.e., 1,640 directions.
Unlike the original setup in \citet{zero-2023}, we follow \citet{nllb, lslora_2024} to standardize the validation and testing processes using Flores\footnote{\url{https://github.com/openlanguagedata/flores}.}, which is a high-quality parallel dataset available for over 200 languages.
Specifically, we use the \textit{dev} and \textit{devtest} sets of Flores for validation and testing, containing 997 and 1,012 sentences per language, respectively.

In the evaluation, we set the beam size to 5 in inference.
We employ four automatic metrics to evaluate inference results on the test set:
(1) spBLEU \cite{nllb}, a variant of BLEU \cite{bleu, sacrebleu} used for Flores, unifies tokenization across languages through an open-source tokenizer\footnote{\url{https://tinyurl.com/flores200sacrebleuspm}.};
(2) chrF++ \cite{chrf, chrf++} assesses character-level overlap and balances precision with recall;
(3) COMET\footnote{All COMET scores are computed using \textit{Unbabel/wmt22-comet-da} \cite{comet-22}.} \cite{comet} evaluates quality by comparing generated translations, reference translations, and source sentences at a representation level;
(4) we report the off-target ratio \cite{massive-2020} as a supplementary metric, because the testing tool\footnote{\url{https://github.com/LlmKira/fast-langdetect}.} is not fully accurate as it relies on recognizing language-specific tokens.
Since COMET and off-target ratio evaluations lack support for certain languages, we compute these scores only for supported languages, as listed in Appendix \ref{appendix:metric}.

\subsection{Configuration and Baseline}\label{section:benchmark_configuration}
The modeling follows the manner of Transformer \cite{transformer} with an embedding size of 1,024, an inner size of 4,096, and 16 attention heads.
We divide the models into two configurations based on model depth.
We first stack 12 layers and balance the number of layers between the encoder and decoder in Enc-dec, resulting in 242M parameters for Enc-dec and 217M for Dec-only.
Then, models include 24 layers in the second configuration, yielding 418M parameters for Enc-dec and 368M for Dec-only.

Apart from the vanilla Enc-dec and Dec-only, we also reproduce four related methods mentioned in Section \ref{section:related} as baselines:
(1) LAVS \cite{target-off} adds language-specific tokens to the shared dictionary;
(2) CL \cite{Constras-2021} aligns sentence-level semantic representations across languages using the encoder output;
(3) LCS \cite{lcs-2024} strengthens translation instructions for Enc-dec by biasing token representations in the encoder with a target language embedding. This mechanism resembles $\ve{r}$ but uses a different operation;
(4) TDO \cite{tdo} strengthens translation instructions for Dec-only by dividing the process of Dec-only into two phases, specifically, encoding source tokens with stronger target language features in the first phase, and then, concatenating the encoded source tokens and target tokens to feed into Dec-only models.
We list all implementation and training details of these baselines in Appendix \ref{appendix:training_ec40}.

\subsection{Result}\label{section:benchmark_result}
Experimental results shown in Table \ref{tab:result_ec40} exhibit consistent trends across both configurations, where our method consistently performs the best.
The most notable improvement is in the off-target ratio, where the metric reduces from 48.40\% to 4.65\% in 12-layer models and from 26.69\% to 3.65\% in 24-layer models.
These results indicate that registering nearly resolves the off-target problem.
Moreover, although our method does not achieve the highest performance in supervised translation, this is not a drawback, because the higher supervised performance of vanilla models is attributed to the overfitting \cite{spurious_2019, DisentPos-2021}.
Additionally, we observe that registering significantly outperforms LAVS.
Based on our discussion of LAVS and its underlying methods in Section \ref{section:related}, we argue that simply adding language-specific parameters is not a cost-efficient solution.

We also observe that the gains of spBLEU scores from related methods tend to diminish as the number of model parameters increases.
In the 12-layer models, the highest gain among four related methods over vanilla models in zero-shot translation is 3.88.
In 24-layer models, the improvement decreases to 2.15.
However, our method achieves more consistent improvements with gains of 5.19 and 3.62 in 12-layer and 24-layer models, respectively.
From this comparison, we can conclude that registering demonstrates superior scalability.

\section{Experiment: Pre-trained Models}\label{section:pretrain}
\begin{table}[t]
  \centering
    \resizebox{0.8\linewidth}{!}{
    \begin{tabular}{lll}
      \toprule
      Family (Group) & Languages & Bridge \\
      \midrule
      English* & \texttt{en} & \texttt{en} \\
      Germanic &  \texttt{de}, \texttt{nl}, \texttt{sv}, \texttt{da}, \texttt{af} & \texttt{de}, \texttt{nl} \\
      Romance & \texttt{fr}, \texttt{es}, \texttt{it}, \texttt{pt}, \texttt{ro} & \texttt{fr}, \texttt{es} \\
      Slavic & \texttt{ru}, \texttt{cs}, \texttt{pl}, \texttt{bg}, \texttt{uk} &  \texttt{ru}, \texttt{cs} \\
      Malayo-Polynesian & \texttt{id}, \texttt{ms}, \texttt{jv}, \texttt{tl} & \texttt{id}, \texttt{ms} \\
      Asian* & \texttt{ja}, \texttt{zh}, \texttt{ko}, \texttt{vi} & \texttt{ja}, \texttt{zh}\\
      \bottomrule 
    \end{tabular}
    }
  \caption{Languages in data collection.
   Languages are shown by their ISO 639-1 codes.
   Decoration with * indicates a language group instead of a language family.}
  \label{tab:collection_languages}
\end{table}
\begin{table*}[!ht]
    \centering
    \resizebox{\textwidth}{!}{
    \begin{tabular}{clcccccccccccccc}
    \toprule
        ~ & ~ & \multicolumn{2}{c}{English} & \multicolumn{2}{c}{Germanic} & \multicolumn{2}{c}{Romance} & \multicolumn{2}{c}{Slavic} & \multicolumn{2}{c}{Mal.-Polyn.} & \multicolumn{2}{c}{Asian} & ~ \\ 
    \midrule
        \multicolumn{2}{c}{Model} & $\rightarrow$ & $\leftarrow$ & $\rightarrow$ & $\leftarrow$ & $\rightarrow$ & $\leftarrow$ & $\rightarrow$ & $\leftarrow$ & $\rightarrow$ & $\leftarrow$ & $\rightarrow$ & $\leftarrow$ & avg. \\ 
    \midrule
        \multirow{3}{*}{M2M} & 483M & 30.36  & 31.92  & 24.40  & 22.58  & 24.01  & 25.81  & 22.59  & 23.40  & 17.94  & 16.50  & 18.30  & 18.37  & 22.10  \\ 
        ~ & 615M & 30.69  & 31.98  & 26.35  & 25.56  & 25.47  & 27.52  & 24.02  & 24.77  & 20.11  & 17.49  & 19.31  & 19.09  & 23.65  \\ 
        ~ & 1.2B & 35.92  & 35.14  & 29.51  & 26.82  & 28.40  & 28.38  & 26.58  & 26.19  & 18.09  & 17.57  & 15.48  & 20.07  & 24.69  \\ 
    \midrule
        \multirow{3}{*}{NLLB} & 615M & 35.85  & 41.04  & 28.13  & 27.41  & 27.46  & 29.09  & 25.40  & 25.33  & 25.39  & 24.35  & 20.72  & 19.42  & 26.05  \\ 
        ~ & 1.3B & 38.08  & 43.42  & 30.52  & 30.17  & 29.63  & 31.42  & 27.84  & 28.25  & 28.08  & 26.87  & 23.50  & 21.06  & 28.51  \\ 
        ~ & 3.3B & 39.80  & \textbf{45.08}  & 31.93  & 31.77  & 30.88  & 32.62  & 29.29  & 30.13  & 29.81  & \textbf{28.08}  & 25.18  & 22.56  & 30.01  \\ 
    \midrule
        \multirow{2}{*}{GPT} & 3.5 turbo & 38.27  & 42.37  & 31.01  & 31.02  & 30.09  & 32.73  & 28.56  & 27.85  & 26.75  & 22.81  & 23.61  & 24.08  & 28.66  \\ 
        ~ & 4o mini & \textbf{41.49}  & 43.97  & 33.09  & 31.92  & 31.40  & 34.03  & 30.54  & 30.69  & \textbf{31.01}  & 27.20  & \textbf{26.34}  & \textbf{27.53}  & 31.09  \\ 
    \midrule
        \multirow{2}{*}{MITRE} & 466M & \better40.20  & 42.60  & \better32.14  & 31.51  & \better31.32  & \better33.26  & \better29.36  & 29.80  & 28.46  & 26.16  & 24.05  & \better23.56  & 29.77  \\ 
        ~ & 913M & \better41.16  & \best44.17  & \best\textbf{33.34}  & \best\textbf{32.95}  & \best\textbf{32.53}  & \best\textbf{34.23}  & \best\textbf{30.74}  & \best\textbf{31.26}  & \better29.90  & \best27.22  & \better25.93  & \better25.58  & \best\textbf{31.15}  \\ 
    \bottomrule
    \end{tabular}
    }
    \caption{Averaged spBLEU scores comparing MITRE with baselines.
    The off-target ratio is not reported due to the near-zero values in these large-scale models.
   chrF++ and COMET scores are provided in Tables \ref{tab:result_large_chrf} and \ref{tab:result_large_comet}, as discussed in Appendix \ref{appendix:results_large}.
   Mal.-Polyn. abbreviates Malayo-Polynesian, and other abbreviations follow Table \ref{tab:result_ec40}.
   Prompts used for GPT are reported in Appendix \ref{appendix:prompts}.
   Additionally, we use \compactbetter{green boxes} to highlight scores exceeding NLLB-3.3B and \compactbest{blue boxes} for those surpassing GPT-4o mini, where \compactbest{blue box} has the priority.}
   \label{tab:result_large}
\end{table*}
\subsection{Data Collection with Bridge Languages}\label{section:pretrain_dataset}
A robust and practical MNMT-specific model requires training across multiple directions rather than English-involved directions only \cite{massive-2020, massive_2022}.
However, collecting data for every possible translation direction is infeasible, as the number of directions grows exponentially with the number of supported languages.
In this work, limited by our computational resources, we adopt the Bridge Language strategy \citep{m2m} to collect data across 24 languages spanning more than five language families.

As shown in Table \ref{tab:collection_languages}, we group languages by family except for English.
The Germanic, Romance, and Slavic groups belong to European language families, and the Malayo-Polynesian differs significantly from these European languages.
In addition, we define a special group, Asian, which includes four languages predominantly spoken in the Asian continent: \texttt{ja}, \texttt{zh}, \texttt{ko}, and \texttt{vi}.
While these languages belong to different families, they share certain similarities due to their geographic proximity.
We designate the two most resource-rich languages in each group as bridge languages and follow these rules for data collection: (1) \texttt{en} connects with all languages; (2) bridge languages connect with each other; (3) bridge languages connect with the remaining languages within their respective groups.
Given that \texttt{ms} cannot meet (2) and (3) due to its low resource, we collect additional data for \texttt{ms}.
Based on the above strategy, out of 552 possible translation directions, we collect data from the reproduced version of the NLLB dataset\footnote{\url{https://opus.nlpl.eu/NLLB/corpus/version/NLLB}} \cite{nllb} for a total of 194 directions, resulting in 9.3 billion sentence pairs for our pre-training.\footnote{Appendix \ref{appendix:pretrain_data} reports the data distribution at the language-family and language level.}

\subsection{Configuration and Baseline}\label{section:pretrain_configure}
We begin by training a vocabulary of 160,000 tokens using SentencePiece \cite{sentencepiece} on 150 million sentences randomly sampled from the training set. 
We then pre-train two models, named MITRE (\textbf{m}ult\textbf{i}lingual \textbf{t}ranslation with \textbf{re}gisters), on 80 V100 GPUs with 466 million and 913 million parameters, respectively.
We report complete details of modeling and training in Appendix \ref{appendix:training_mitre}.
The validation and testing process aligns with Section \ref{section:benchmark_dataset}.

We compare our model against not only state-of-the-art MNMT-specific models, but also commercial LLMs, because commercial LLMs present the upper limit of fine-tuning open-sourced LLMs into MNMT-specific models (Section \ref{section:related}).
First, the MNMT models include three versions of M2M (483M\footnote{The official name is M2M-418M, however, this model actually has 483M parameters.}, 615M, and 1.2B) \cite{m2m} and three versions of NLLB (615M-distilled, 1.3B, and 3.3B) \cite{nllb}.
Also, we include commercial LLMs, GPT-3.5 Turbo\footnote{Version is \textit{gpt-3.5-turbo-0125}.} \cite{gpt3} and GPT-4o mini\footnote{Version is \textit{gpt-4o-mini-2024-07-18}.} \cite{gpt4}.
Meanwhile, we include these NLLB models as baselines\footnote{Due to the limitation of computational resources, NLLB-3.3B is not included in fine-tuning with full parameters.} in our fine-tuning experiments.
Specifically, we create three scenarios randomly selecting 5, 25, and 100 translation directions from the possible directions.
Then, we perform fine-tuning with full parameters and with LoRA \cite{lora-2022} on the Flores \textit{dev}, which contains 997 sentence pairs per direction.
Fine-tuning settings are provided in Appendix \ref{appendix:training_finetune}.

\subsection{Main Results}\label{section:pretrain_result}
The experimental results comparing MITRE with baselines are shown in Table \ref{tab:result_large}.
We observe that although NLLB-3.3B surpasses MITRE-913M by 0.91 spBLEU points for translations into English and by 0.86 points for Malayo-Polynesian languages, MITRE-913M consistently achieves higher scores in other translation directions, with an overall average gain of 1.14 points. 
Given that NLLB even surpasses GPT-4o mini by 1.11 points in English translation, we infer that MITRE, a Dec-only model with registering, demonstrates better generalization than NLLB, based on Enc-dec.
Notably, scaling parameters of NLLB from 1.3B to 3.3B yields only a gain of 1.50 points, while MITRE attains a comparable gain of 1.38 points with an additional 450M parameters.
Furthermore, the alignment of training and validation loss for two MITRE models (Appendix \ref{appendix:training_mitre}) reinforces our conclusion in Section \ref{section:benchmark_result} that registering provides superior scalability.
Finally, based on all experimental results, we conclude that MITRE-466M performs competitively with NLLB-3.3B, while MITRE-913M not only outperforms NLLB-3.3B but also competes with GPT-4o mini\footnote{Table \ref{tab:result_large_comet} shows that the COMET scores of MITRE-913M are lower than GPT-4o mini. Therefore, although spBLEU and chrF++ scores of MITRE-913M are higher, we only claim that MITRE-913M performs competitively with GPT-4o mini. We provide an additional discussion in Appendix \ref{appendix:results_large}.}, showing the practical potential of our models.

\begin{table}[t]
    \centering
    \resizebox{\linewidth}{!}{
    \begin{tabular}{lccccccc}
    \toprule
        ~  & ~ & \multicolumn{2}{c}{5-direction} & \multicolumn{2}{c}{25-direction} & \multicolumn{2}{c}{100-direction} \\ 
    \midrule
        model & ~ & spB. & com. & spB. & com. & spB. & com. \\ 
    \midrule
        \multirow{3}{*}{N.-615M} & pre. & 24.00  & 82.91  & 25.88  & 83.71  & 25.37  & 83.35  \\ 
        ~  & lora & 24.68 & 83.23 & 26.84 & 84.10 & 26.41 & 84.00  \\
        ~  & f.t. & 25.59  & 83.80  & 27.32  & 84.29  & 26.70  & 84.17  \\ 
        \cdashline{2-8}
        \multirow{3}{*}{N.-1.3B} & pre. & 26.59  & 84.86  & 28.33  & 85.41  & 27.82  & 85.17  \\ 
        ~  & lora & 27.39 & 85.20 & 29.49 & 85.87 & 29.18 & 85.87  \\ 
        ~ & f.t. & 28.50  & 85.78  & 30.13  & 86.09  & 29.61  & 86.05  \\
        \cdashline{2-8}
        \multirow{2}{*}{N.-3.3B} & pre. & 27.95 & 85.60 & 29.70 & 86.16 & 29.31 & 85.98  \\ 
        ~ & lora &  29.05 & 86.08 & 31.23 & 86.63 & 30.98 & 86.71  \\ 
        \cdashline{2-8}
        \multirow{3}{*}{M.-466M} & pre. & 24.51  & 83.73  & 28.71  & 85.41  & 29.07  & 85.26  \\ 
        ~  & lora & 26.37 & 84.47 & 29.97 & 86.09 & 30.42 & 86.27  \\ 
        ~  & f.t. & 28.19  & 85.28  & 30.61  & 86.36  & 30.81  & 86.45  \\ 
        \cdashline{2-8}
        \multirow{3}{*}{M.-913M} & pre. & 25.52  & 84.56  & 29.95  & 86.07  & 30.37  & 85.92  \\ 
        ~  & lora & \better28.14 & \better85.59 & \better31.68 & \better86.90 & \better32.33 & \better87.15  \\ 
        ~  & f.t. & \best\textbf{30.09}  & \best\textbf{86.45}  & \best\textbf{32.47} & \best\textbf{87.23} & \best\textbf{32.73}  & \best\textbf{87.35}  \\ 
    \bottomrule
    \end{tabular}
    }
    \caption{
    Averaged spBLEU and COMET scores of results on three fine-tuning scenarios, where the specific translation directions are listed in Appendix \ref{appendix:training_finetune}.
    N., M., pre., and f.t. abbreviate NLLB, MITRE, pre-trained models, and fine-tuning with full parameters, respectively.
    The best score is in bold, \compactbest{blue boxes} highlights the largest gain in f.t. relative to pre., and \compactbetter{green boxes} highlights the largest gain in lora.
    }
    \label{tab:finetune}
\end{table}

\subsection{Fine-tuning Results}\label{section:finetune}
Table \ref{tab:finetune} shows the fine-tuning results. By comparing NLLB and MITRE, we observe that MITRE outperforms NLLB in both scenarios: fine-tuning on a few translation directions and fine-tuning on multiple translation directions simultaneously. Specifically, we find that performance gains from fine-tuning increase with model size, and our MITRE-913M shows the highest improvement in both full parameter fine-tuning and LoRA-based fine-tuning.
Additionally, since pre-trained models of both NLLB and MITRE achieve near-zero off-target ratios, these gains can be attributed to increased quality rather than addressing the off-target problem. This suggests that MITRE has a higher performance ceiling, likely due to our cost-effective data collection strategy, which may have constrained MITRE from reaching its theoretical maximum. Therefore, due to MITRE's superior fine-tuning capability, we reaffirm that its practical potential is remarkable.

\begin{table}[t]
    \centering
    \resizebox{0.9\linewidth}{!}{
    \begin{tabular}{ccccccc}
    \toprule
        \#layer & register & mask & spB.$\uparrow$ & chrf$\uparrow$ & com.$\uparrow$ & off.$\downarrow$ \\ 
    \midrule
        \multirow{3}{*}{12} & \redcross & \redcross & 8.34  & 22.34  & 55.71  & 22.21  \\ 
        ~ & \greencheck & \redcross & 8.19  & 22.96  & 55.72  & 32.11  \\ 
        ~ & \greencheck & \greencheck & 11.92  & 29.02  & 61.19  & 4.65  \\ 
    \midrule
        \multirow{3}{*}{24} & \redcross & \redcross & 10.82  & 26.04  & 59.95  & 19.01  \\ 
        ~ & \greencheck & \redcross & 8.91  & 24.06  & 57.60  & 34.22  \\ 
        ~ & \greencheck & \greencheck & 13.12  & 30.51  & 63.54  & 3.65  \\
    \bottomrule
    \end{tabular}
    }
    \caption{Averaged scores of ablation study on EC-40.
    Here, register means adding registers, and mask means modifying the attention mask.}
    \label{tab:ablation}
    \vspace{-0.5em}
\end{table}
\begin{figure}[t]
    \centering
        \includegraphics[width=0.9\linewidth]{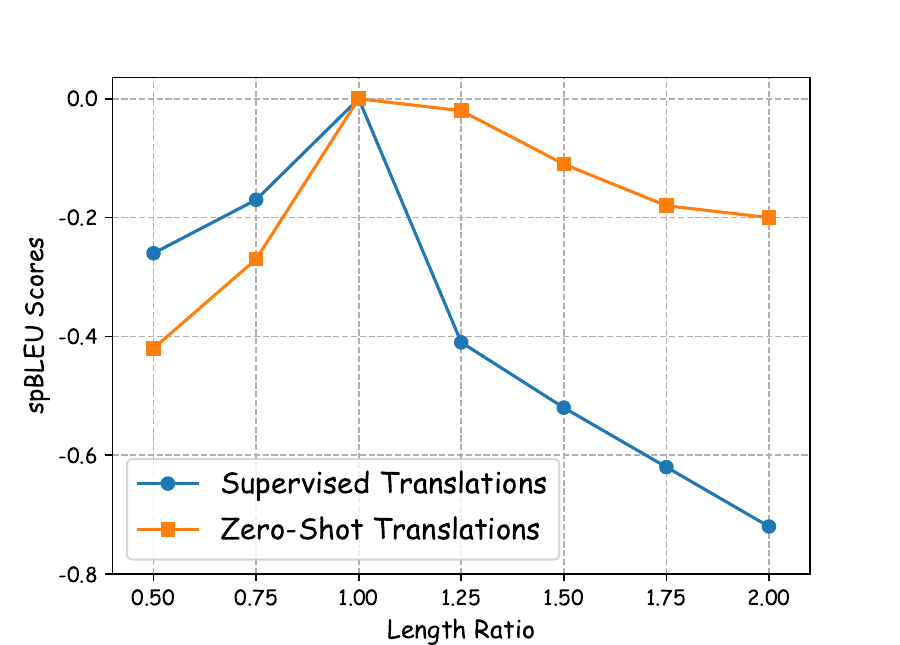}
    \caption{The spBLEU score variations on EC-40 where the x-axis is $\text{len}(\ve{x}')/\text{len}(\ve{r})$, where only the length of $\ve{r}$ is changed and $\ve{x}'$ is fixed.}
    \label{fig:ablation}
\end{figure}

\section{Discussion}
\subsection{Ablation Study}
We conduct two ablation studies to measure the impact of registering.
First, we decompose registering into two steps: (1) adding registers and (2) modifying the attention mask. As shown in Table \ref{tab:ablation}, merely adding registers reduces the performance of vanilla Dec-only models; registering only becomes effective after modifying the attention mask. This result aligns with expectations, i.e., the model without constraints on generation defaults to relying directly on source tokens instead of registers.

In Section \ref{section:registering}, we state that the lengths of $\ve{r}$ and $\ve{x}$ are matched to ensure a one-to-one correspondence between registers and source tokens.
To validate this design, we vary the length of $\ve{r}$ while keep $\ve{x}'$ fixed to observe performance trends.
Specifically, a ratio less than 1.0 means that $\ve{r}$ is an augmenting of $\ve{x}'$, a ratio greater than 1.0 means $\ve{r}$ is a compressing of $\ve{x}'$, and a ratio of 1.0 means the registering.
The trend illustrated in Figure \ref{fig:ablation} empirically supports that registering is the optimal mechanism.

\begin{table}[t]
    \centering
    \resizebox{0.95\linewidth}{!}{
    \begin{tabular}{clcccc}
    \toprule
        Ratio & Mechanism & T.1 S. & T.2 S. & Dist. & Entropy$\uparrow$ \\
        \midrule
        0.75 & augmenting     & 1.68 & 0.72 & 2.67 & 5.15 \\
        1    & registering    & 1.80 & 0.78 & 2.09 & \textbf{5.25} \\
        1.25 & compressing & 2.13 & 0.94 & 1.53 & 4.91 \\
        1.5  & compressing & 2.15 & 0.94 & 1.62 & 4.73 \\
    \bottomrule
    \end{tabular}
    }
    \caption{
    Attention mechanism across different ratios, i.e., $\text{len}(\ve{x})/\text{len}(\ve{r})$.
    T.1 S. and T.2 S. denote average top-1 and top-2 attention scores; Dist. is the average positional distance between the top-2 source tokens; Entropy measures the diversity of source token selection.
    Higher entropy indicates more diverse attention; lower values suggest focus on a narrower set of tokens.
    }
    \label{tab:attention_mechanism}
\end{table}
To analyze the mechanism differences among augmenting, registering, and compressing, we analyze the attention alignment between registers and source tokens based on models trained on EC-40.
For each register, we extract the source tokens with the highest and the second-highest attention scores.
We then measure the entropy of the source token selection distribution for each sentence, i.e., how frequently each source token is selected as the top-1 attention.
Results based on 100 random instances are summarized in Table \ref{tab:attention_mechanism}.
Among the three mechanisms, registering yields the highest entropy, indicating a more diverse use of source tokens, i.e., each token is likely to be selected as a top attention target at least once.
Augmenting, by contrast, results in lower scores and longer distances, suggesting redundancy and diffused attention.
Compressing, while producing high scores, focuses on short contiguous spans and exhibits reduced entropy, implying potential neglect of broader contextual information.
These findings supplement the empirical results in Figure \ref{fig:ablation} and reinforce registering as the optimal mechanism.

\begin{figure}[t]
    \includegraphics[width=0.9\linewidth]{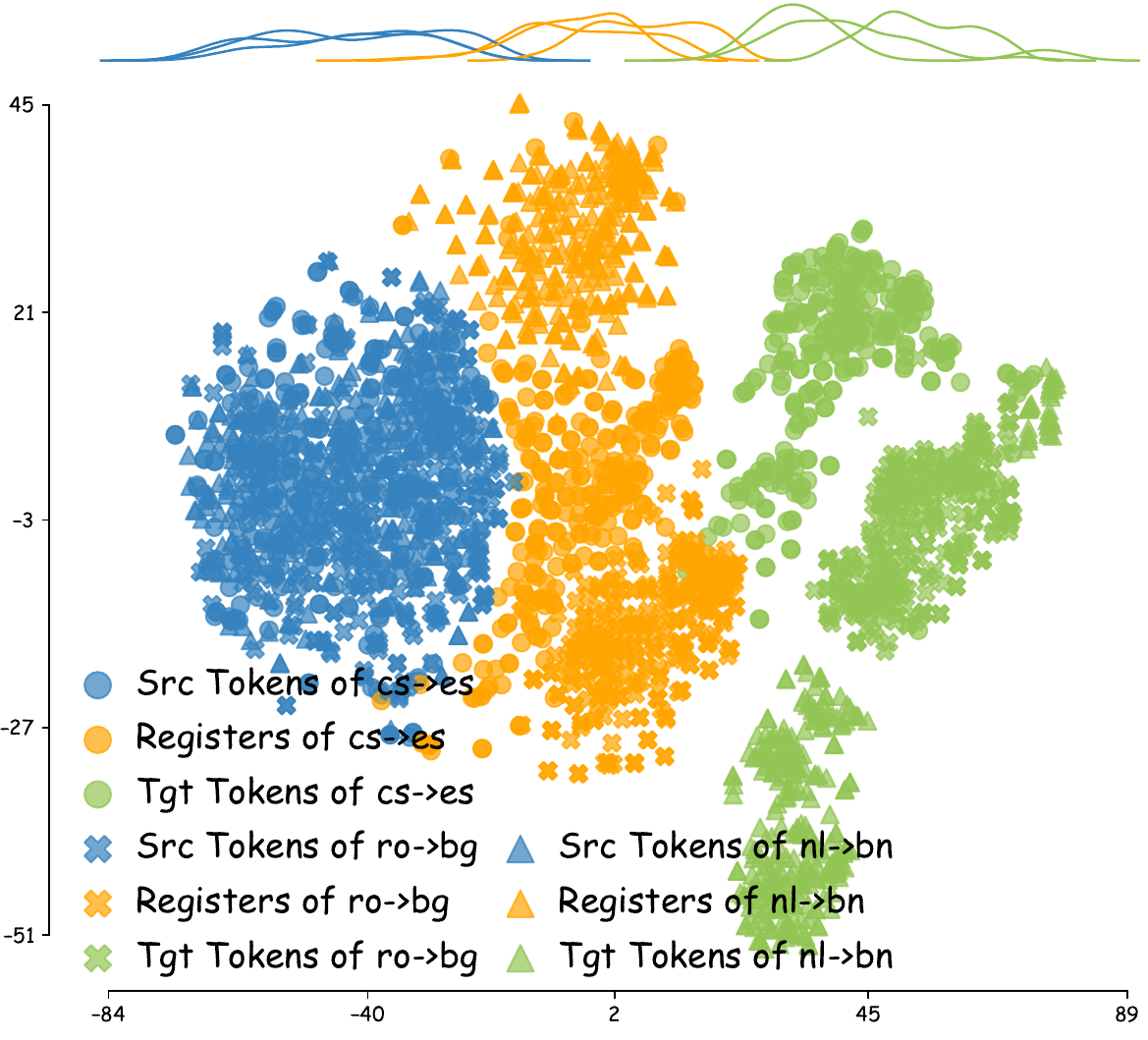}
    \caption{
    2D distribution of token-level representations extracted from the output of the 24th layer of a model trained on EC-40.
    Each class listed in the legend contains 300 randomly sampled tokens.
    Appendix \ref{appendix:rep} shows the representational distributions from other layers.
    }
    \label{fig:representation}
\end{figure}
\begin{figure*}[t]
    \centering
    \begin{subfigure}[t]{0.15\linewidth}
        \includegraphics[width=\textwidth]{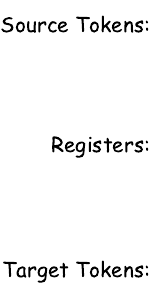}
    \end{subfigure}
    \begin{subfigure}[t]{0.4\linewidth}
        \includegraphics[width=\linewidth]{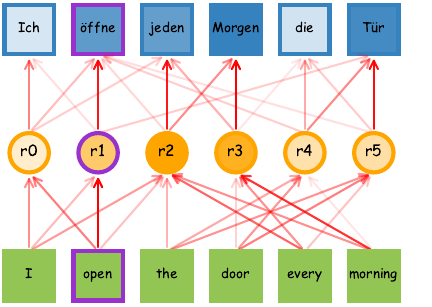}
        \caption{Inseparable Verb}
        \label{fig:inseparable}
    \end{subfigure}
    \begin{subfigure}[t]{0.4\linewidth}
        \includegraphics[width=\linewidth]{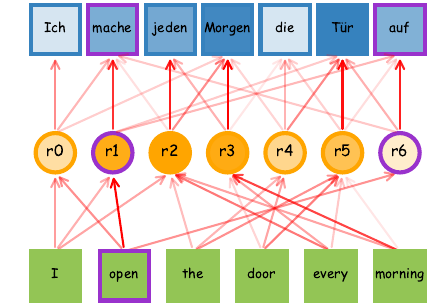}
        \caption{Separable Verb}
        \label{fig:separable}
    \end{subfigure}
    \caption{
    Token-level attention weights illustration, where the weight of each token is averaged across all heads of a model trained on EC-40.
    \ref{fig:inseparable} and \ref{fig:separable} illustrate two instances translated from \texttt{de} to \texttt{en}. The top-3 attention directions for each token are labeled, with darker colors indicating higher attention weights. Note that while the target tokens for these two instances are identical, their source tokens are not, because the verbs in \ref{fig:inseparable} and \ref{fig:separable} are semantically equivalent but have different forms. To aid understanding, we highlight the verbs with purple borders: ``\"{o}ffne'' in \ref{fig:inseparable} and ``mache ... auf'' in \ref{fig:separable} both correspond to the target verb ``open''. Then, the registers with the highest attention weights associated with these verbs are also marked with purple borders.
    }
    \label{fig:attention_analysis}
\end{figure*}
\subsection{Mechanism: registering source tokens to target language spaces}\label{section:analysis}
We first reveal the registering mechanism at the token level from two perspectives: (1) representations of registers are located in the intended target language space, and (2) registers carry the semantics of the positionally-aligned source token.

We analyze token representations by randomly selecting 100 translation instances from three translation directions and applying t-SNE \cite{tsne} to reduce them to two dimensions.
Figure \ref{fig:representation} shows the representation distribution in the final layer.
First, source tokens, registers, and target tokens are clearly separated into distinct spaces, indicating that the model can distinguish their different functions.
Additionally, source token representations from different languages cluster, suggesting that the model processes them in a language-agnostic manner. Most importantly, for registers and target tokens, token representations for the three translation directions cluster in separate spaces. This supports our design, where the register representation is located in the intended target language space.

The relationship between source tokens and registers can be exhibited by analyzing the attention weights in generation based on a simple and interesting grammar of \texttt{de}.
Figure \ref{fig:attention_analysis} shows two translation instances translated from \texttt{de} to \texttt{en}, which have the same semantics.
We observe that the target tokens in both examples are identical, while their source tokens differ only in the verb form (highlighted with a purple border). Specifically, in Figure \ref{fig:inseparable}, the verb is a single token, ``\"{o}ffne'', whereas in Figure \ref{fig:separable}, it consists of two distant tokens, ``mache ... auf''. Despite this difference, ``\"{o}ffne'' and ``mache ... auf'' share the same semantics and can be replaced by each other.
Figure \ref{fig:attention_analysis} presents the attention weights for each token representation.
In Figure \ref{fig:inseparable}, the highest attention weight of the verb in the target tokens, i.e., ``open'', comes from $r_1$; in Figure \ref{fig:separable}, ``open" pays the highest attention weight to $r_1$ and $r_6$.
Meanwhile, $r_1$ in Figure \ref{fig:inseparable} aligns positionally with ``\"{o}ffne'', and in Figure \ref{fig:separable}, $r_1$ and $r_6$ align positionally with ``mache ... auf''.
Additionally, we observe that the highest attention weight of a register always comes from its positionally-aligned source token\footnote{Appendix \ref{appendix:attn} shows more examples in Russian, Chinese, and Japanese, where all instances follow this statement.}.

\begin{figure}[t]
    \centering
        \includegraphics[width=\linewidth]{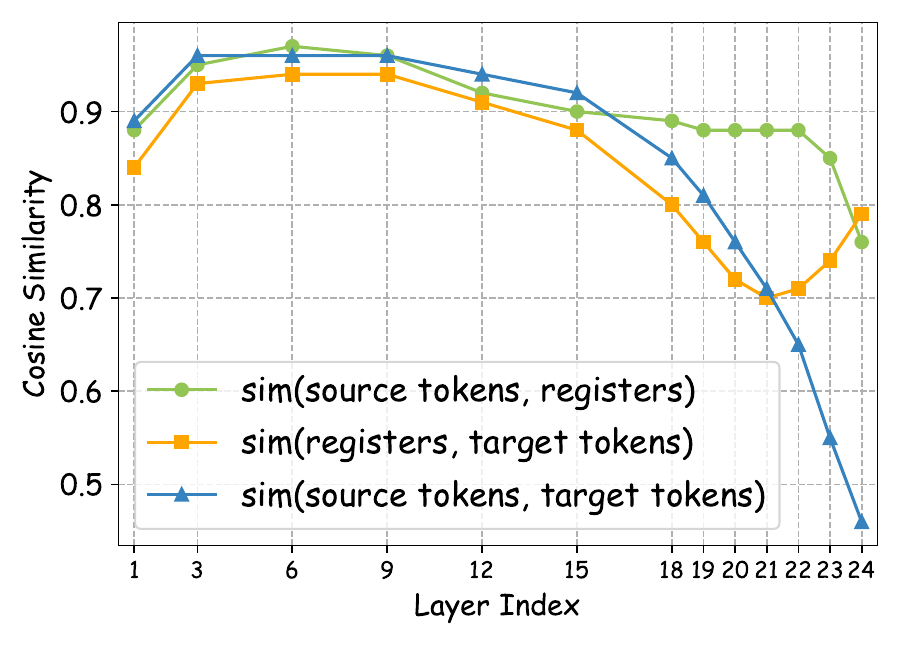}
    \caption{Layer-wise cosine similarity among sentence-level representations of source tokens, registers, and target tokens. 
    Results are averaged over 10 random instances processed by a model trained on EC-40.}
    \label{fig:layer_similarity}
\end{figure}
Beyond theorizing the mechanism through token-level analyses, we also conduct a sentence-level representation analysis\footnote{We follow \citet{DisentPos-2021} to apply mean-pooling over the token representations to obtain sentence representations, and then compute cosine similarity between them. For each translation instance, we extract representations for source tokens, registers, and target tokens.} to validate the mechanism.
As shown in Figure \ref{fig:layer_similarity}, sentence-level cosine similarity between source and target is the highest in the lower layers, reflecting strong semantic alignment due to the use of gold translations and indicating that lower layers primarily encode shared semantic content.
As depth increases, source-target similarity drops noticeably, suggesting that upper layers capture more language-specific features.
Meanwhile, source-register similarity remains relatively stable until the top layers, where it drops sharply.
In contrast, register-target similarity gradually decreases in the middle layers but rises sharply in the top layers.
In the final layer, registers are most similar to the target, followed by those to the source, while source-target similarity is lowest.
These trends indicate that registers transition toward the target space while maintaining semantic ties to the source, supporting our token-level analyses.
Based on the above, we conclude that the register's activation represents the target language and carries the semantics of the positionally-aligned token, namely, registers act as “rethinking” the source from the perspective of the target language.

\section{Conclusion}
In this work, we present registering to address the off-target problem in MNMT-specific models.
By introducing registers and modifying the attention mask, our method ensures that the generation of target tokens depends solely on the activation of registers.
Analytical experiments demonstrate that the activation of registers carries the semantics of source tokens within the target language spaces.
Using this method, we develop and open-source two MNMT-specific models, MITRE-466M and MITRE-913M, supporting translation across 24 languages.
Experimental results show that MITRE performs competitively with commercial LLMs, setting a new state-of-the-art in MNMT.

\section*{Limitations}
A key concern is our limited computational resources.
Given that the training of MITRE-913M has already required 80 Tesla V100 GPUs for one month, MITRE supports 24 languages, and we cannot further increase the supported languages.
Although this number is far greater than the latest research in the community, e.g., 10 languages of \citet{tower_llm_2024} and 5 languages of \citet{ftllm}, this number is fewer than the number of supported languages of M2M, NLLB, and commercial LLMs.
However, the comparison in Section \ref{section:pretrain} is relatively fair.
Specifically (using NLLB as an example), first, our training data is collected from the reproduced version of the NLLB dataset, which includes fewer samples per translation direction than those used for training NLLB models.
Second, as described in Section \ref{section:pretrain_dataset}, our Bridge Language strategy results in fewer supervised translation directions, whereas NLLB is trained on as many directions as possible.
Moreover, NLLB incorporates additional engineering strategies, e.g., back-translation \cite{back_translation} and distillation \cite{online_distill}, whereas MITRE only iterates over the training set.
Also, we directly compare MITRE-466M and MITRE-913M to NLLB-3.3B, where the parameter size difference helps offset the disparity in supported languages.
Finally, we conduct fine-tuning experiments to compare MITRE and NLLB with the same settings.

Another limitation of our approach is the additional computational cost introduced by registers, as they double the number of source tokens.
Based on our measurements on EC-40 using a Tesla V100 GPU, the training time for models with registers is 1.34 times that of the vanilla decoder-only model, 1.63 times that of the vanilla encoder-decoder model, and approximately equivalent (1.01 times) to the previous state-of-the-art method, CL \cite{Constras-2021}.
At inference time, thanks to the KV cache, the model with registers incurs only a linear and affordable increase in inference cost, comparable to other MNMT-specific models.
To validate this, we randomly translate 100 sentences using publicly available implementations of M2M, NLLB, and MITRE from HuggingFace (with batch size 1 and beam size 5), and repeat each experiment 10 times.
Results in Table \ref{tab:generation_cost} confirm our claim.
Additionally, in practical usage of MITRE, the inference cost is substantially lower than that of LLMs due to the smaller number of parameters.

\section*{Ethical Considerations}
Although our training data is collected from public datasets, MITRE has not been evaluated for toxicity or has undergone detoxification.
Thus, while we open-source MITRE, we recommend its use primarily for research purposes or in applications only after thorough appropriate processing.

\begin{table}[t]
    \centering
    \resizebox{\linewidth}{!}{
    \begin{tabular}{lccc}
    \toprule
        Model & \#Tokens & Times (s) & Times per Token (s)\\
        \midrule
        MITRE-466M&3667&64.95&0.0177 \\
        MITRE-913M&3626&85.65&0.0236 \\
        M2M-483M&4226&52.22&0.0124 \\
        M2M-1.2B&4234&89.75&0.0212 \\
        NLLB-600M&3990&41.29&0.0103 \\
        NLLB-1.3B&4104&74.37&0.0181 \\
        NLLB-3.3B&3837&82.15&0.0214 \\
    \bottomrule
    \end{tabular}
    }
    \caption{
    Generation cost measurement.
    Tokens refer to the total tokens generated during inference, and Times refer to the time cost of inference, which is counted by seconds.
    }
    \label{tab:generation_cost}
\end{table}

\bibliography{custom}
\clearpage

\appendix
\section{Description of EC-40}\label{appendix:ec40}
EC-40 is an English-centric dataset introduced by \citet{zero-2023}. In addition to English, it includes 40 languages spanning five language families, with each family containing eight languages.
These languages are categorized into four tiers based on data availability: High, Medium, Low, and Extra Low. Each non-English language is paired with English, resulting in 80 supervised translation directions used for training and 1,560 zero-shot translation directions.
Details of this dataset are summarized in Table \ref{tab:ec40}.

\section{Description of Pre-training Dataset}\label{appendix:pretrain_data}
\begin{figure*}[t]
   \centering
      \begin{subfigure}[b]{0.55\linewidth}
        \includegraphics[width=\linewidth]{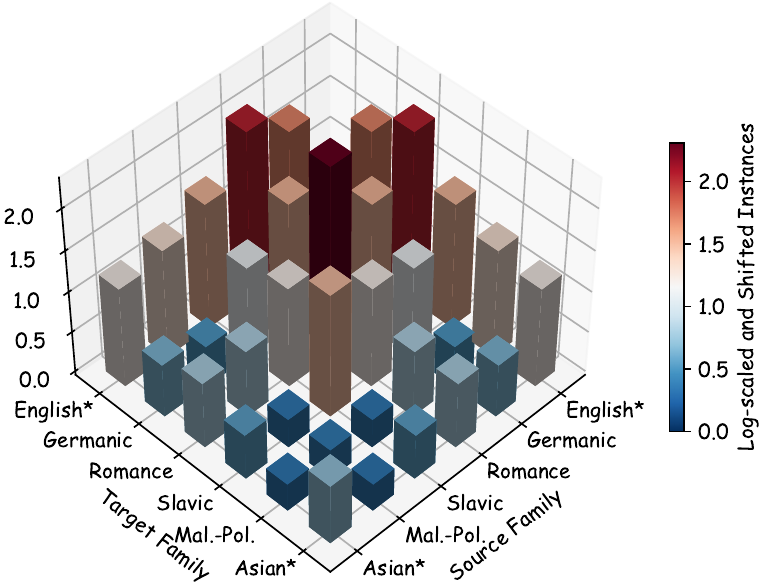}
        \caption{Family Level}
        \label{fig:family_level}
      \end{subfigure}
      \begin{subfigure}[b]{0.43\linewidth}
        \centering
        \includegraphics[width=\linewidth]{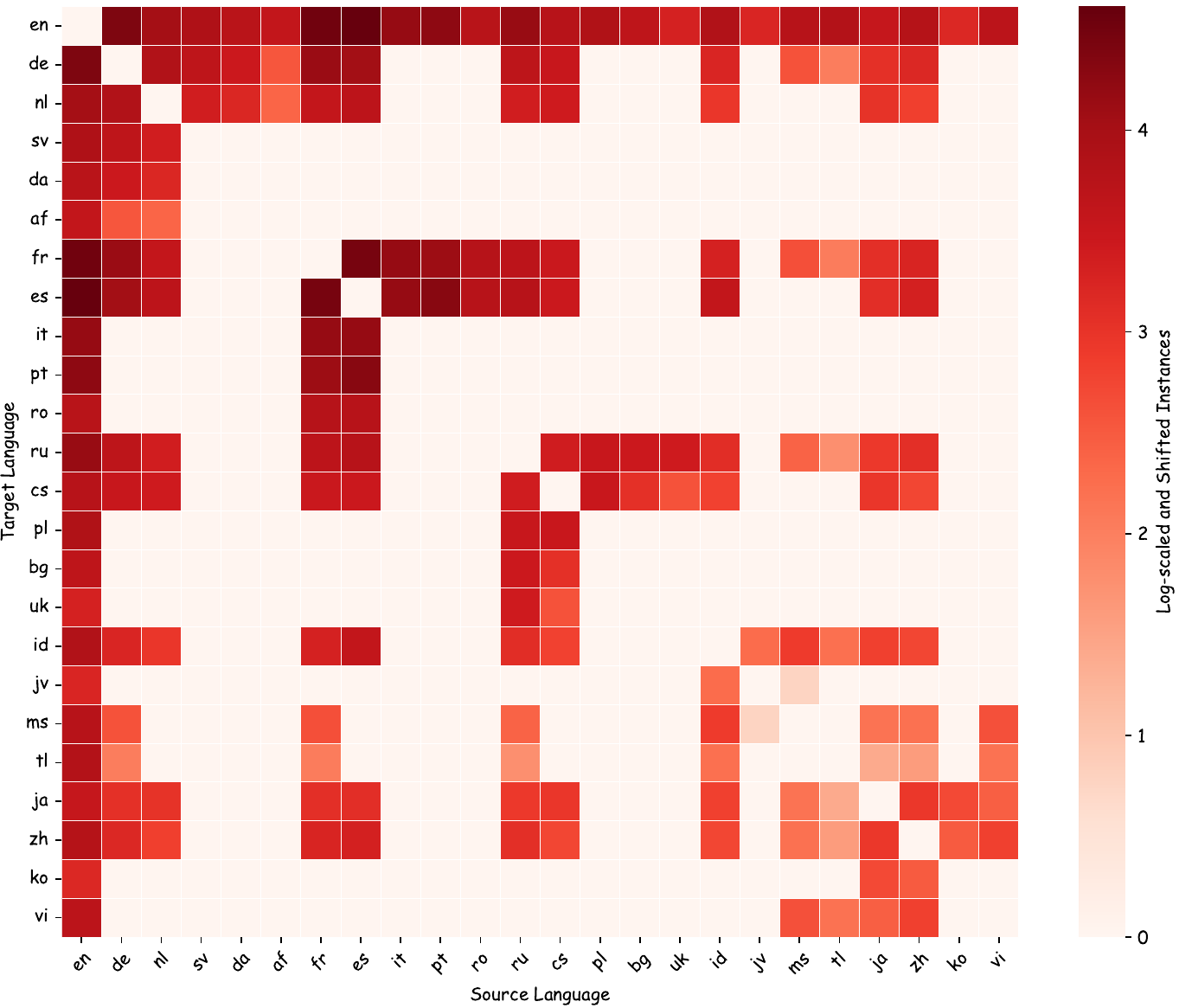}
        \caption{Language Level}
        \label{fig:language_level}
      \end{subfigure}
    \caption{
    Data distribution of our pre-training dataset. Figure \ref{fig:family_level} shows data size distribution at the family level, while Figure \ref{fig:language_level} displays data size at the language level. In \ref{fig:family_level}, non-zero values are scaled by log10 and adjusted by subtracting 7 for clearer visualization. In \ref{fig:language_level}, non-zero values are also scaled by log10 and shifted by subtracting the minimum value to enhance illustration clarity.
    }
    \label{fig:data_distribution}
\end{figure*}

Our pre-training dataset comprises 24 languages, as detailed in Table \ref{tab:dataset}.
As described in Section \ref{section:pretrain_dataset}, our data collection strategy results in 9.3 billion translation instances across 194 translation directions.
The data distribution is visualized at the family level in Figure \ref{fig:family_level} and at the language level in Figure \ref{fig:language_level}. Additionally, Figure \ref{fig:language_level} highlights which translation directions are supervised and which are zero-shot. Notably, translation directions involving \texttt{ms} are also indicated in Figure \ref{fig:language_level}.

\section{Training Details of EC-40}\label{appendix:training_ec40}
\paragraph{Training configurations}
We employ Fairseq \cite{fairseq}, an open-source toolkit, to implement our models with methods mentioned in Section \ref{section:benchmark_configuration}.
First, we directly reuse the vocabulary and binary training data provided by \citet{target-off}\footnote{\url{https://github.com/Smu-Tan/ZS-NMT-Variations}}.
Note that we include only supervised translation directions in validation.
We train on 8 Tesla V100 GPUs, setting \textit{memory-efficient-fp16} in Fairseq, with a maximum input of 2048 source tokens per GPU and a gradient accumulation of 16 steps.
Both input and output token lengths are limited to 256, and we share the embedding layer between the encoder and decoder.
We use a seed of 1234, a learning rate of 0.0005 with the inverse square root schedule and a warmup of 4000 steps, the Adam optimizer \cite{adam}, dropout of 0.1, attention dropout of 0.1, a label smoothing rate of 0.1, no weight decay, and the temperature sampling with $T=5$ \cite{temperatureSample}.
Finally, we train for 200,000 steps, averaging the last 5 checkpoints, saving by epoch.

\paragraph{Configurations of Related Methods}
Generally, both our method and these related methods share the same hyper-parameters as the vanilla models.
However, there are some method-specific configurations we have to notice.
(1) For LAVS \cite{target-off}, we directly reuse their code\footnote{\url{https://github.com/PKUnlp-icler/Off-Target-MNMT}} and add around 12k language-specific tokens into the shared vocabulary, resulting in 12.8M additional parameters in modeling;
(2) For CL \cite{Constras-2021}, we directly reuse their code\footnote{\url{https://github.com/PANXiao1994/mRASP2}} and set the contrastive learning temperature to 0.1, which is the optimal setting according to their reports;
(3) For LCS \cite{lcs-2024}, the model follows another translation instruction strategy of \citet{m2m} by adding a source language tag at the beginning of the source tokens and a target language tag at the beginning of the target tokens.
we re-implement their code and, in the case of 12-layer models where the encoder has 6 layers, apply LCS biasing at the 5th encoder layer; For models with 12 encoder layers, we apply it at the 8th encoder layer;
(4) For TDO \cite{tdo}, we also reuse their code\footnote{\url{https://github.com/zhiqu22/PhasedDecoder}} and, based on their ablation study, set the number of layers for the first stage to 3 in 12-layer models and to 6 in 24-layer models to allow stronger zero-shot translation ability.
\begin{table}[t]
    \centering
    \resizebox{0.9\linewidth}{!}{
    \begin{tabular}{lcccc}
    \toprule
        Family & ISO code & Flores code & Language & script \\ 
    \midrule
        \multirow{6}{*}{Germanic} & en & eng\_Latn & English & Latin \\ 
        ~ & de & deu\_Latn & German & Latin \\ 
        ~ & nl & nld\_Latn & Dutch & Latin \\ 
        ~ & sv & swe\_Latn & Swedish & Latin \\ 
        ~ & da & dan\_Latn & Danish & Latin \\ 
        ~ & af & afr\_Latn & Afrikaans & Latin \\
    \midrule
        \multirow{5}{*}{Romance} & fr & fra\_Latn & French & Latin \\ 
        ~ & es & spa\_Latn & Spanish & Latin \\ 
        ~ & it & ita\_Latn & Italian & Latin \\ 
        ~ & pt & por\_Latn & Portuguese & Latin \\ 
        ~ & ro & ron\_Latn & Romanian & Latin \\
    \midrule
        \multirow{5}{*}{Slavic} & ru & rus\_Cyrl & Russian & Cyrillic \\ 
        ~ & cs & ces\_Latn & Czech & Latin \\ 
        ~ & pl & pol\_Latn & Polish & Latin \\ 
        ~ & bg & bul\_Cyrl & Bulgarian & Cyrillic \\ 
        ~ & uk & ukr\_Cyrl & Ukrainian & Cyrillic \\
    \midrule
        \multirow{4}{*}{\begin{minipage}{0.3cm}Malayo-Polynesian\end{minipage}} & id & ind\_Latn & Indonesian & Latin \\ 
        ~ & ms & zsm\_Latn & Malay & Latin \\ 
        ~ & jv & jav\_Latn & Javanese & Latin \\ 
        ~ & tl & fil\_Latn & Filipino & Latin \\
    \midrule
       \multirow{4}{*}{Asian*} & ko & kor\_Hang & Korean & Hangul \\ 
        ~ & vi & vie\_Latn & Vietnamese & Latin \\ 
        ~ & ja & jpn\_Jpan & Japanese & Kanji; Kana \\ 
        ~ & zh & cmn\_Hans & Chinese & Chinese \\
    \bottomrule
    \end{tabular}
    }
    \caption{
    Details of the dataset in our pre-training.
    The decoration * on Asian means a group instead of a language family.
    We not only list the ISO 630-1 code for each language but also list the Flores code to help search corresponding resources from Flores+.
    }
    \label{tab:dataset}
\end{table}

\begin{figure}[t]
    \centering
        \includegraphics[width=\linewidth]{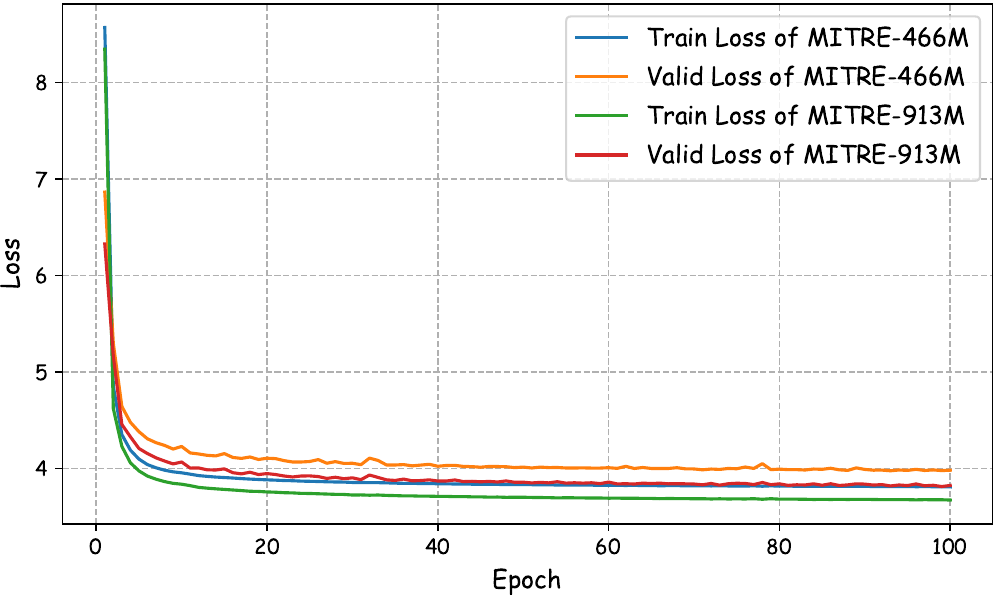}
    \caption{The training and validation loss in pre-training MITRE.
    We report the first 100 epochs, each with approximately 2262 steps.
    }
    \label{fig:loss}
\end{figure}
\section{Training Details of MITRE}\label{appendix:training_mitre}
We employ Fairseq to implement MITRE mentioned in Section \ref{section:pretrain_configure}, and two versions of MITRE have the different configurations in modeling and have the same configuration in training.
Specifically, MITRE-466M is configured with an embedding size of 1,024, an inner size of 4,096, 16 attention heads, and 24 layers.
MITRE-913M, a larger model with expanded width and depth, has an embedding size of 1,280, an inner size of 5,120, 20 heads, and 36 layers.
In training, we first train a shared SentencePiece vocabulary \cite{sentencepiece} with a size of 160,000 by 150 million sentences randomly sampled from the training set.
We include only supervised translation directions in validation.
Then, we train on 80 Tesla V100 GPUs, setting \textit{memory-efficient-fp16} in Fairseq, with a maximum input of 1408 source tokens per GPU and a gradient accumulation of 10 steps.
In practice, this setup results in each batch containing approximately 0.91 million source tokens.
Given the large batch size, we set the learning rate of 0.002 with the inverse square root schedule and the warmup of 8000 steps.
We also use a seed of 42, the Adam optimizer \cite{adam}, dropout of 0.1, attention dropout of 0.1, a label smoothing rate of 0.1, no weight decay, and the temperature sampling with $T=1$ \cite{temperatureSample}.
We train for 300,000 steps and save a checkpoint per 10,000 steps.
Finally, we average the last 5 checkpoints.
Figure \ref{fig:loss} shows the variations of training and validation loss.
We can observe that the trends of MITRE-466M and MITRE-913M are highly consistent.

\section{Training Details of Fine-tuning}\label{appendix:training_finetune}
\paragraph{Selecting Directions}
We use \textit{random.sample} in Python to randomly select translation directions for fine-tuning, setting the seed to 0.
We define three scenarios, including 5, 25, and 100 translation directions.
It is important to note that \textit{random.sample} causes the 5 and 25 directions to be subsets of the 100 directions.
Specifically, the first 5 and first 25 directions in the 100-direction set correspond to the other two scenarios.
The 100 directions are:
\texttt{jv}{\small$\rightarrow$}\texttt{sv},
\texttt{ms}{\small$\rightarrow$}\texttt{id}, \texttt{de}{\small$\rightarrow$}\texttt{tl}, \texttt{ru}{\small$\rightarrow$}\texttt{pl}, \texttt{ko}{\small$\rightarrow$}\texttt{jv}, \texttt{zh}{\small$\rightarrow$}\texttt{bg}, \texttt{ms}{\small$\rightarrow$}\texttt{en}, \texttt{pl}{\small$\rightarrow$}\texttt{ru}, \texttt{zh}{\small$\rightarrow$}\texttt{af}, \texttt{uk}{\small$\rightarrow$}\texttt{ko}, \texttt{pt}{\small$\rightarrow$}\texttt{jv}, \texttt{ko}{\small$\rightarrow$}\texttt{ro}, \texttt{fr}{\small$\rightarrow$}\texttt{da}, \texttt{cs}{\small$\rightarrow$}\texttt{pl}, \texttt{fr}{\small$\rightarrow$}\texttt{af}, \texttt{da}{\small$\rightarrow$}\texttt{fr}, \texttt{ru}{\small$\rightarrow$}\texttt{sv}, \texttt{fr}{\small$\rightarrow$}\texttt{pl}, \texttt{pl}{\small$\rightarrow$}\texttt{tl}, \texttt{da}{\small$\rightarrow$}\texttt{ro}, \texttt{sv}{\small$\rightarrow$}\texttt{es}, \texttt{bg}{\small$\rightarrow$}\texttt{jv}, \texttt{zh}{\small$\rightarrow$}\texttt{en}, \texttt{da}{\small$\rightarrow$}\texttt{cs}, \texttt{uk}{\small$\rightarrow$}\texttt{ms}, \texttt{tl}{\small$\rightarrow$}\texttt{es}, \texttt{bg}{\small$\rightarrow$}\texttt{de}, \texttt{pt}{\small$\rightarrow$}\texttt{nl}, \texttt{vi}{\small$\rightarrow$}\texttt{bg}, \texttt{tl}{\small$\rightarrow$}\texttt{id}, \texttt{ru}{\small$\rightarrow$}\texttt{bg}, \texttt{nl}{\small$\rightarrow$}\texttt{ms}, \texttt{en}{\small$\rightarrow$}\texttt{uk}, \texttt{da}{\small$\rightarrow$}\texttt{sv}, \texttt{jv}{\small$\rightarrow$}\texttt{ms}, \texttt{en}{\small$\rightarrow$}\texttt{nl}, \texttt{zh}{\small$\rightarrow$}\texttt{vi}, \texttt{bg}{\small$\rightarrow$}\texttt{ja}, \texttt{ro}{\small$\rightarrow$}\texttt{ja}, \texttt{bg}{\small$\rightarrow$}\texttt{ru}, \texttt{nl}{\small$\rightarrow$}\texttt{tl}, \texttt{vi}{\small$\rightarrow$}\texttt{es}, \texttt{ja}{\small$\rightarrow$}\texttt{pt}, \texttt{cs}{\small$\rightarrow$}\texttt{uk}, \texttt{da}{\small$\rightarrow$}\texttt{ko}, \texttt{af}{\small$\rightarrow$}\texttt{it}, \texttt{jv}{\small$\rightarrow$}\texttt{zh}, \texttt{zh}{\small$\rightarrow$}\texttt{cs}, \texttt{sv}{\small$\rightarrow$}\texttt{da}, \texttt{ko}{\small$\rightarrow$}\texttt{pt}, \texttt{cs}{\small$\rightarrow$}\texttt{nl}, \texttt{pt}{\small$\rightarrow$}\texttt{vi}, \texttt{nl}{\small$\rightarrow$}\texttt{en}, \texttt{vi}{\small$\rightarrow$}\texttt{ja}, \texttt{es}{\small$\rightarrow$}\texttt{nl}, \texttt{tl}{\small$\rightarrow$}\texttt{ru}, \texttt{ru}{\small$\rightarrow$}\texttt{es}, \texttt{ja}{\small$\rightarrow$}\texttt{jv}, \texttt{ro}{\small$\rightarrow$}\texttt{zh}, \texttt{nl}{\small$\rightarrow$}\texttt{ro}, \texttt{fr}{\small$\rightarrow$}\texttt{jv}, \texttt{cs}{\small$\rightarrow$}\texttt{fr}, \texttt{fr}{\small$\rightarrow$}\texttt{cs}, \texttt{uk}{\small$\rightarrow$}\texttt{jv}, \texttt{ko}{\small$\rightarrow$}\texttt{bg}, \texttt{cs}{\small$\rightarrow$}\texttt{da}, \texttt{es}{\small$\rightarrow$}\texttt{ro}, \texttt{ms}{\small$\rightarrow$}\texttt{sv}, \texttt{ja}{\small$\rightarrow$}\texttt{cs}, \texttt{cs}{\small$\rightarrow$}\texttt{en}, \texttt{da}{\small$\rightarrow$}\texttt{pl}, \texttt{jv}{\small$\rightarrow$}\texttt{tl}, \texttt{pl}{\small$\rightarrow$}\texttt{pt}, \texttt{zh}{\small$\rightarrow$}\texttt{sv}, \texttt{pl}{\small$\rightarrow$}\texttt{de}, \texttt{fr}{\small$\rightarrow$}\texttt{ro}, \texttt{pt}{\small$\rightarrow$}\texttt{zh}, \texttt{zh}{\small$\rightarrow$}\texttt{id}, \texttt{pl}{\small$\rightarrow$}\texttt{fr}, \texttt{ko}{\small$\rightarrow$}\texttt{ru}, \texttt{it}{\small$\rightarrow$}\texttt{bg}, \texttt{es}{\small$\rightarrow$}\texttt{de}, \texttt{cs}{\small$\rightarrow$}\texttt{tl}, \texttt{af}{\small$\rightarrow$}\texttt{pt}, \texttt{fr}{\small$\rightarrow$}\texttt{ru}, \texttt{da}{\small$\rightarrow$}\texttt{nl}, \texttt{da}{\small$\rightarrow$}\texttt{af}, \texttt{ms}{\small$\rightarrow$}\texttt{fr}, \texttt{ko}{\small$\rightarrow$}\texttt{cs}, \texttt{en}{\small$\rightarrow$}\texttt{jv}, \texttt{pl}{\small$\rightarrow$}\texttt{uk}, \texttt{bg}{\small$\rightarrow$}\texttt{uk}, \texttt{af}{\small$\rightarrow$}\texttt{tl}, \texttt{ro}{\small$\rightarrow$}\texttt{bg}, \texttt{de}{\small$\rightarrow$}\texttt{pl}, \texttt{de}{\small$\rightarrow$}\texttt{vi}, \texttt{uk}{\small$\rightarrow$}\texttt{nl}, \texttt{id}{\small$\rightarrow$}\texttt{ja}, \texttt{nl}{\small$\rightarrow$}\texttt{zh}, \texttt{zh}{\small$\rightarrow$}\texttt{pl}

\paragraph{Fine-tuning Configurations}
All fine-tuning experiments use the same settings.
We conduct experiments on 8 Tesla V100 GPUs, with a maximum of 1024 tokens per GPU and a gradient accumulation of 2 steps.
Based on the pre-trained model, we set the learning rate to 0.0001 with a warmup step of 1 (for launching the inverse square root schedule), and train for 10 epochs.
Finally, we use the last epoch for testing.

\paragraph{LoRA Configurations}
We adopt the setting of \citet{lora-2022} to implement LoRA components for pre-trained models. Specifically, LoRA is only implemented for Query and Value in the attention mechanism with a rank of 8.
As a result, the learnable parameters of NLLB series are 1.18M, 2.36M, and 4.72M, respectively, and the learnable parameters of MITRE series are 0.78M and 1.47M, respectively.

\section{Prompts for GPT}\label{appendix:prompts}
Our prompts for GPT series follow: Translating the following sentence from [SRC] to [TGT]: [INPUT].
Here, [SRC] and [TGT] are the source and target language names following Table \ref{tab:dataset}, and [INPUT] is the source sentence.
We find that GPT occasionally repeats [INPUT] in the output.
Once it happens, we manually remove the [INPUT] before evaluation.

\section{Details of Evaluation Metrics}\label{appendix:metric}
In evaluating the performance of models trained on EC-40, some languages lack support from COMET (\textit{Unbabel/wmt22-comet-da}) and the off-target ratio (\textit{fast-langdetect}).
Notably, \textit{fast-langdetect} operates by word recognition, so we also exclude certain supported languages that exhibit low recognition success rates.
We list the supported languages in this section.
\paragraph{Languages in COMET:}
\texttt{en}, \texttt{bg}, \texttt{so}, \texttt{ca}, \texttt{da}, \texttt{be}, \texttt{bs}, \texttt{es}, \texttt{uk}, \texttt{am}, \texttt{hi}, \texttt{ro}, \texttt{no}, \texttt{de}, \texttt{cs}, \texttt{pt}, \texttt{nl}, \texttt{mr}, \texttt{is}, \texttt{ne}, \texttt{ur}, \texttt{ha}, \texttt{sv}, \texttt{gu}, \texttt{ar}, \texttt{fr}, \texttt{ru}, \texttt{it}, \texttt{pl}, \texttt{sr}, \texttt{sd}, \texttt{he}, \texttt{af}, \texttt{kn}, \texttt{bn}.

\paragraph{Languages in Off-target Ratio:}
\texttt{en}, \texttt{bg}, \texttt{da}, \texttt{es}, \texttt{uk}, \texttt{hi}, \texttt{ro}, \texttt{de}, \texttt{cs}, \texttt{pt}, \texttt{nl}, \texttt{mr}, \texttt{ur}, \texttt{sv}, \texttt{gu}, \texttt{ar}, \texttt{fr}, \texttt{ru}, \texttt{it}, \texttt{pl}, \texttt{he}, \texttt{kn}, \texttt{bn}, \texttt{be}, \texttt{mt}, \texttt{am}, \texttt{is}, \texttt{sd}.

\section{Supplementary Results of EC-40}\label{appendix:results_ec40}
In Section \ref{section:benchmark_result} and Table \ref{tab:result_ec40}, we report spBLEU scores and off-target ratio.
In this appendix, we report chrF++ and COMET scores in Tables \ref{tab:result_ec40_chrf} and \ref{tab:result_ec40_comet}, respectively.
Overall, four metrics show consistent trends across this benchmark.

\begin{figure}[t]
   \centering
      \begin{subfigure}[b]{0.8\linewidth}
        \includegraphics[width=\linewidth]{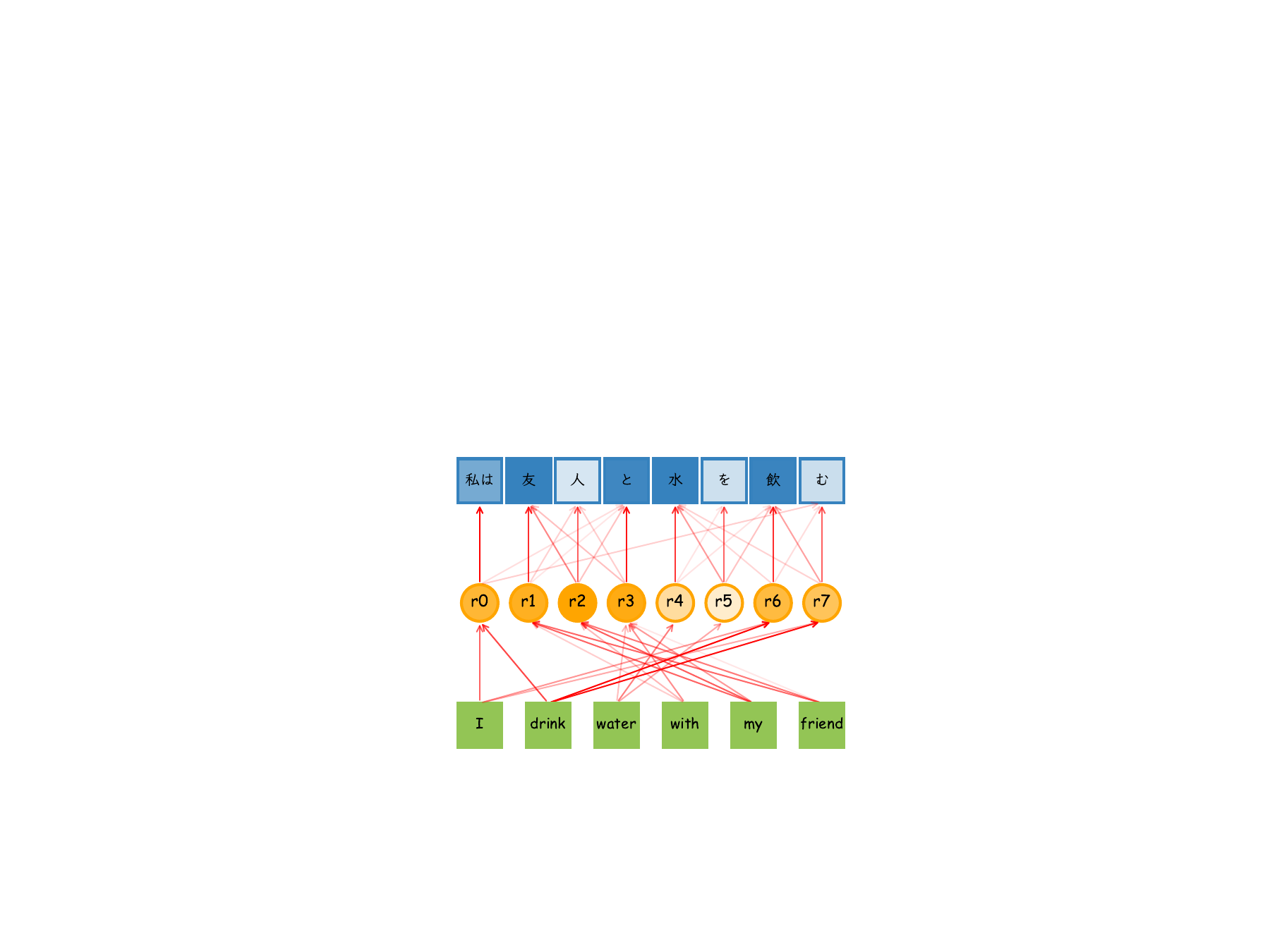}
        \caption{\texttt{ja}$\rightarrow$\texttt{en}}
        \label{fig:ja_attention}
      \end{subfigure}
      \begin{subfigure}[b]{0.8\linewidth}
        \centering
        \includegraphics[width=\linewidth]{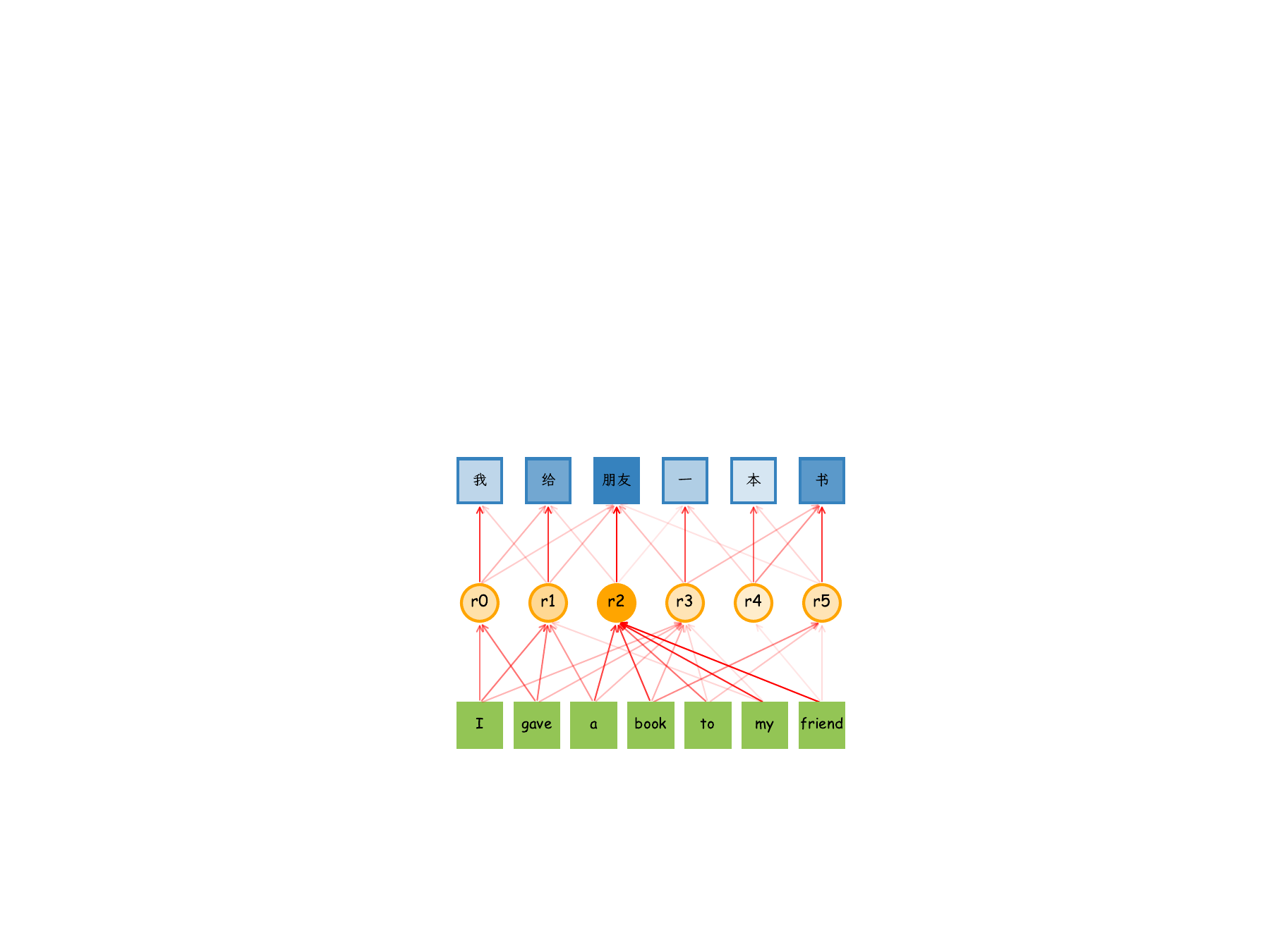}
        \caption{\texttt{zh}$\rightarrow$\texttt{en}}
        \label{fig:zh_attention}
      \end{subfigure}
      \begin{subfigure}[b]{0.8\linewidth}
        \centering
        \includegraphics[width=\linewidth]{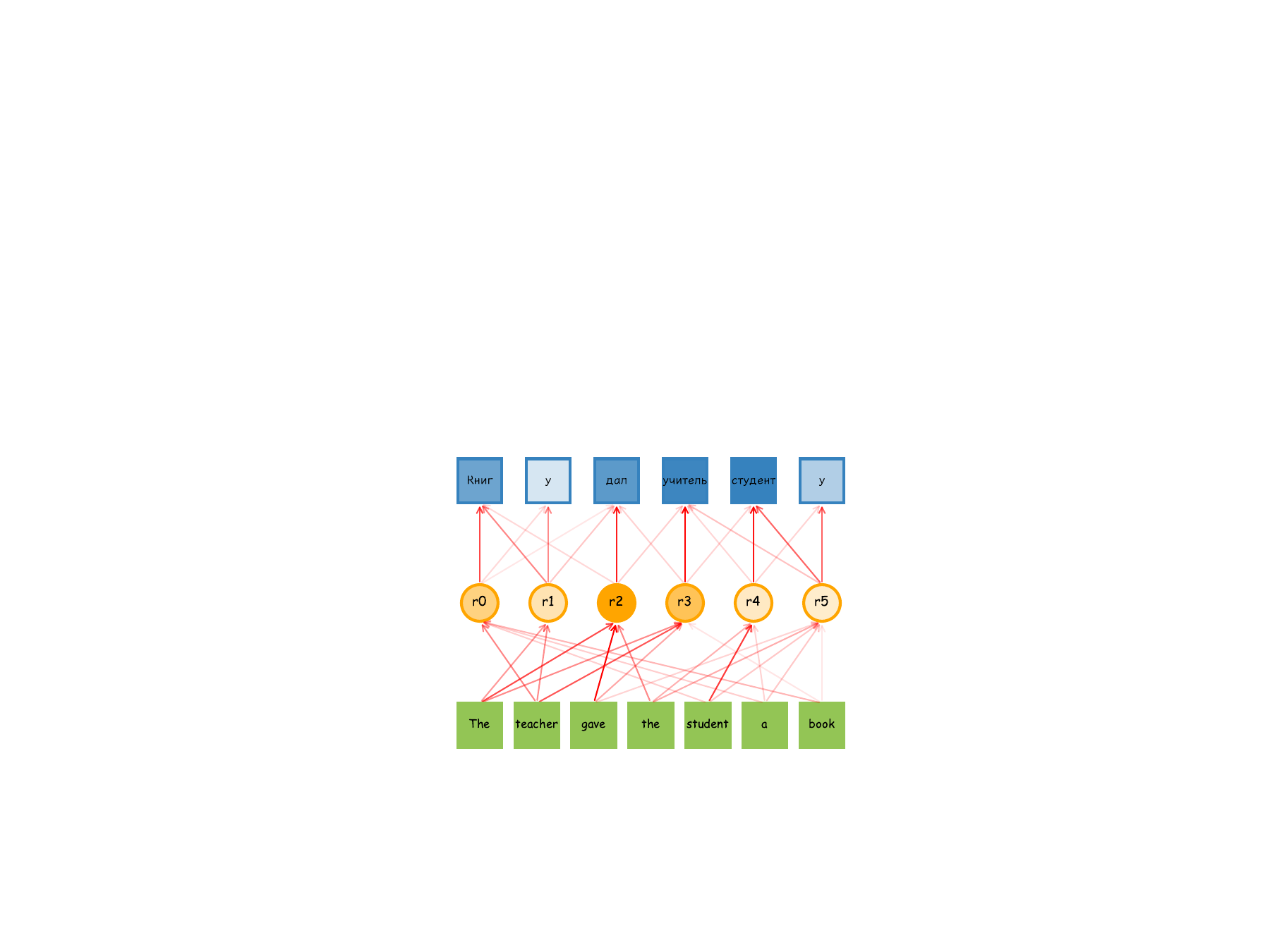}
        \caption{\texttt{ru}$\rightarrow$\texttt{en}}
        \label{fig:ru_attention}
      \end{subfigure}
    \caption{
    Three cases of attention analysis on MITRE-466M.
    Details of this illustration, e.g. colors, classes, and arrows, follow that of Figure \ref{fig:attention_analysis}.
    }
    \label{fig:attention_more_cases}
\end{figure}
\section{Supplementary Results of MITRE}\label{appendix:results_large}
In Section \ref{section:pretrain_result} and Table \ref{tab:result_large}, we report spBLEU scores and do not report the off-target ratio, because the values are near zero across those large-scale pre-trained models.
In this appendix, we report chrF++ and COMET scores in Tables \ref{tab:result_large_chrf} and \ref{tab:result_large_comet}, respectively.
However, when comparing MITRE and commercial LLMs, COMET reveals a different trend: MITRE-913M underperforms GPT-4o mini, despite similar trends in spBLEU and chrF++ scores, which show MITRE-913M as superior to GPT-4o mini.
This suggests that while MITRE generates more accurate sequences relative to the test set, GPT-4o mini produces more fluent and natural output.
This is expected, as commercial models are aligned with human-like styles \cite{align-2023}, whereas MITRE follows the training data's style.
Further supporting this, GPT-4o mini shows a significant improvement over GPT-3.5 turbo (Table \ref{tab:result_large_comet}).
Therefore, based on the results across all three metrics, our claim is not that MITRE-913M outperforms GPT-4o mini, but rather that MITRE-913M outperforms NLLB 3.3B and can compete with GPT-4o mini.

\section{Supporting Experiments for Comparing Enc-dec and Dec-only}\label{appendix:compare}
In Section \ref{section:registering}, we explain our reason for implementing registering in Dec-only. Specifically, Dec-only offers better parameter efficiency than Enc-dec, where the encoder learns the representation of input tokens while the decoder learns the generated tokens (as illustrated in Figure \ref{fig:attention_direction}, where encoded source tokens are shown in gray). In contrast, Dec-only utilizes all parameters for both encoding and generation. Based on this, we employ Dec-only as the backbone of our implementation.

To further support our statement, we provide supporting experiments on EC40, following the experimental setup described in Section \ref{section:benchmark_configuration}.
As shown in Table \ref{tab:result_compare}, registering also significantly improves the performance of Enc-dec. However, the gains are more pronounced in Dec-only, suggesting that registering is particularly well-suited for this architecture.

Additionally, we propose an insight beyond the scope of this work. In Section \ref{section:registering}, we mention a potential cause of the off-target problem: the dilution of translation instruction attention by other tokens \cite{spurious_2019,tdo}. This theory may explain why Dec-only tends to underperform Enc-dec in MNMT \cite{decoderonlymt, decoderonly_2022}, as the effectiveness of $l_{\ve{y}}$ is further diluted by both source and target tokens in the attention mechanism. The results in Table \ref{tab:result_compare} further support this explanation, as registering achieves a much larger performance gain in Dec-only than in Enc-dec.
\begin{table}[t]
    \centering
    \resizebox{\linewidth}{!}{
    \begin{tabular}{lccccccc}
    \toprule
        Methods & \#layer & spBLEU & chrF++ & COMET & Off-target(\%) \\ 
    \midrule
        Enc-dec & \multirow{4}{*}{12} & 6.99 & 19.68 & 53.72 & 48.4 \\ 
        \hspace{0.4em}+registering &  & 10.55 & 27.29 & 58.07 & 7.5 \\ 
        Dec-only &  & 8.34 & 22.34 & 55.71 & 22.21 \\ 
        \hspace{0.4em}+registering &  & \textbf{11.92} & \textbf{29.02} & \textbf{61.19} & \textbf{4.65} \\
    \midrule
        Enc-dec & \multirow{4}{*}{24} & 9.69 & 23.95 & 58.08 & 26.69 \\ 
        \hspace{0.4em}+registering & & 11.13 & 27.91 & 58.67 & 3.91 \\ 
        Dec-only & & 10.82 & 26.04 & 59.95 & 19.01 \\ 
        \hspace{0.4em}+registering & & \textbf{13.12} & \textbf{30.51} & \textbf{63.54} & \textbf{3.65} \\
    \bottomrule
    \end{tabular}
    }
    \caption{
    Results on EC-40.
    For convenience, the score in this table is averaged from all 1,640 translation directions.
    Enc-dec and Dec-only indicate the vanilla models without registering.
    The best score is in bold.
    }
    \label{tab:result_compare}
\end{table}

\section{Supplementary Analysis of Attention}\label{appendix:attn}
To further support our analysis in Section \ref{section:analysis}, we examine additional cases in MITRE-466M where the source and target sentences exhibit significant structural differences.
In all cases, the attention relationship between registers and source tokens remains consistent, i.e., one-to-one attention weights being the most prominent.
Next, we observe the following patterns:
(1) As shown in Figure \ref{fig:ja_attention}, in Japanese, the attention for ``drink'' points to $ r_6 $, while ``friend'' points to $ r_1 $ and $ r_2 $.
(2) As shown in Figure \ref{fig:zh_attention}, ``friend'' points to $ r_2 $, and ``book'' points to $ r_5 $.
(3) As shown in Figure \ref{fig:ru_attention}, ``book'' points to $ r_0 $, and ``student'' points to $ r_4 $.
Given that the attention weights between registers and target tokens highlight the structural differences between source and target sentences, we can state again that registers mirror the corresponding source tokens.

\section{Supplementary Analysis of Representation}\label{appendix:rep}

\begin{figure*}[t]
   \centering
      \begin{subfigure}[b]{0.48\linewidth}
        \includegraphics[width=\linewidth]{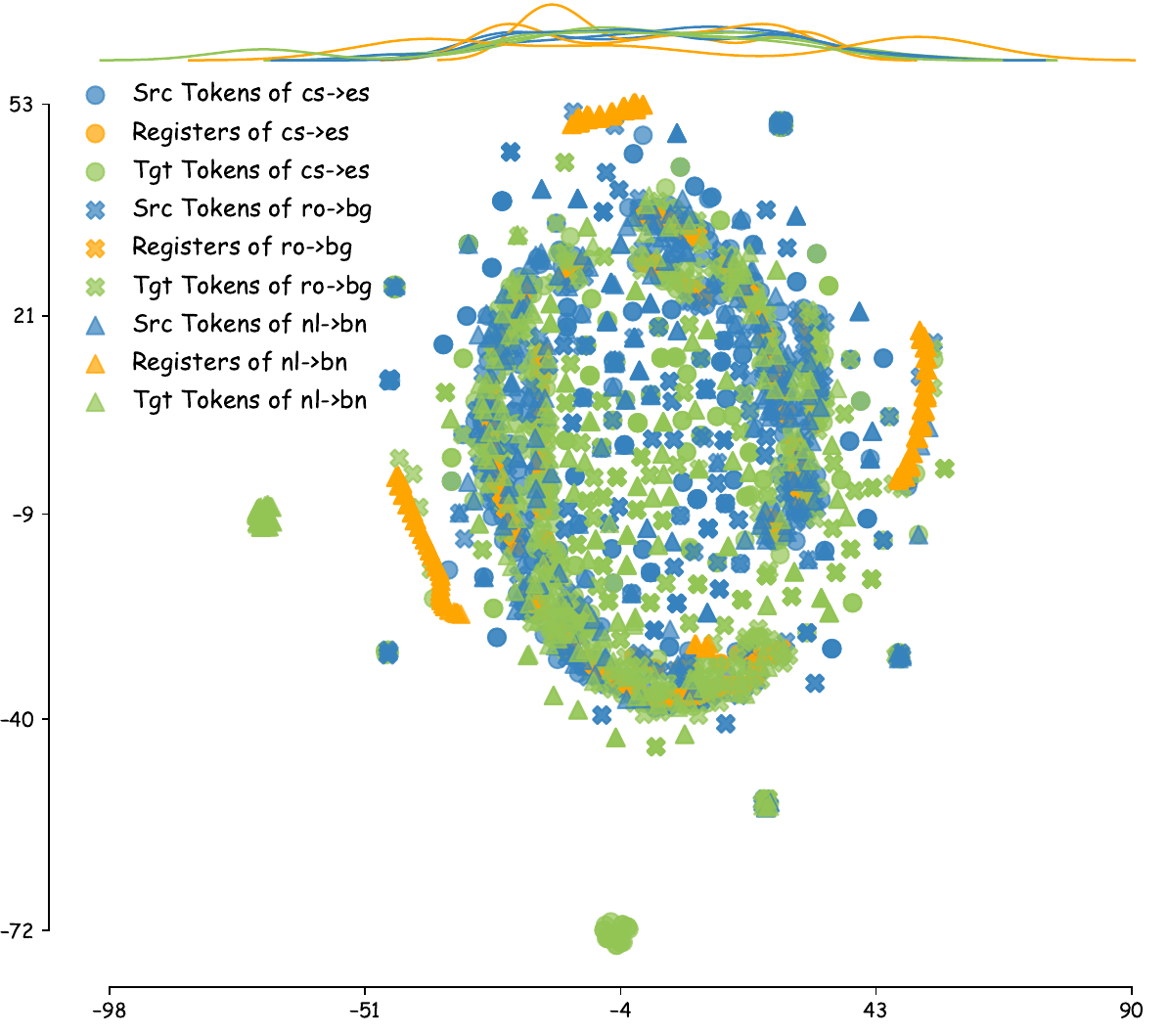}
        \caption{Embedding Layer}
        \label{fig:more_representation_1}
      \end{subfigure}
      \begin{subfigure}[b]{0.48\linewidth}
        \centering
        \includegraphics[width=\linewidth]{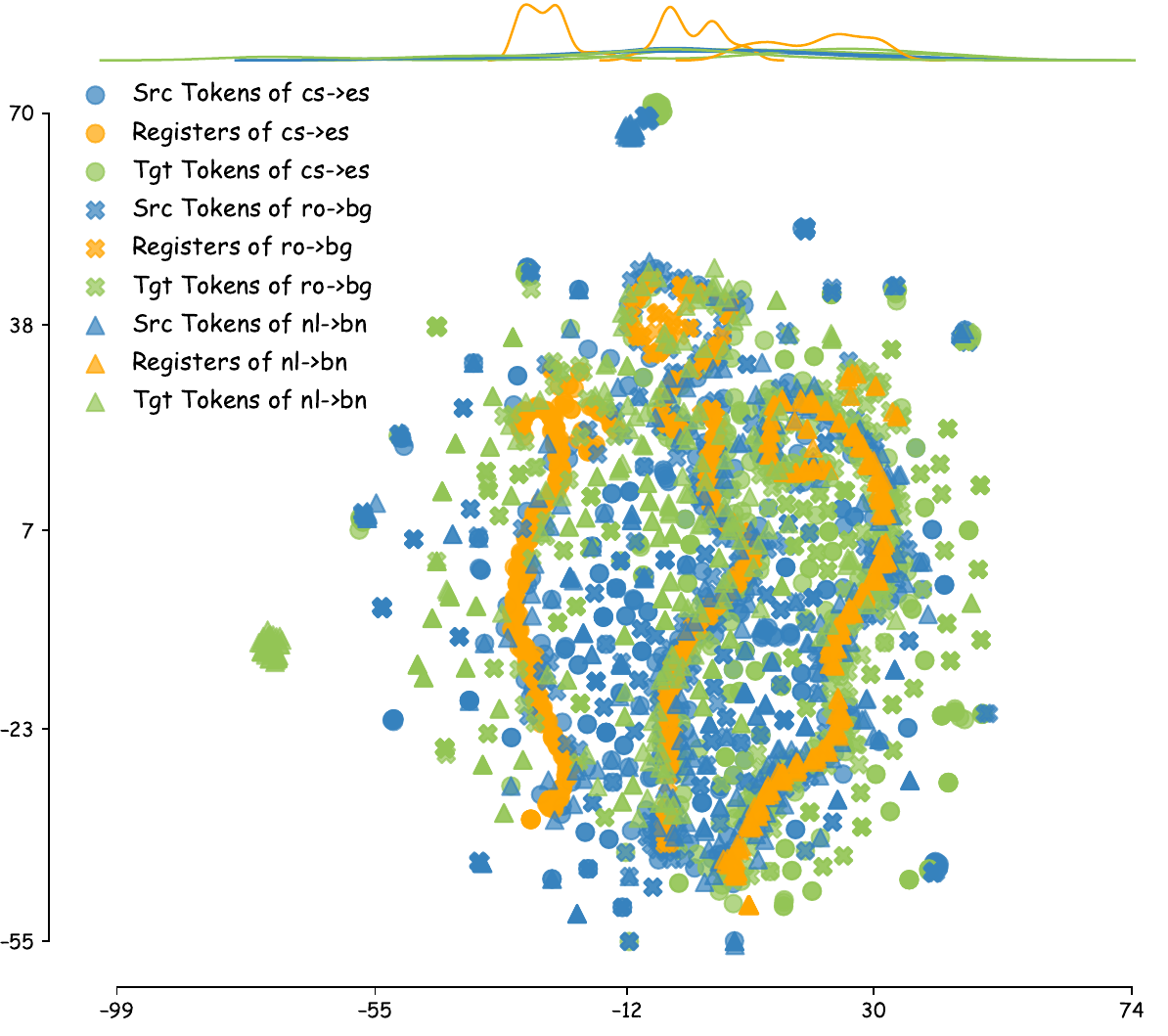}
        \caption{1st Layer}
        \label{fig:more_representation_2}
      \end{subfigure}
      \begin{subfigure}[b]{0.48\linewidth}
        \centering
        \includegraphics[width=\linewidth]{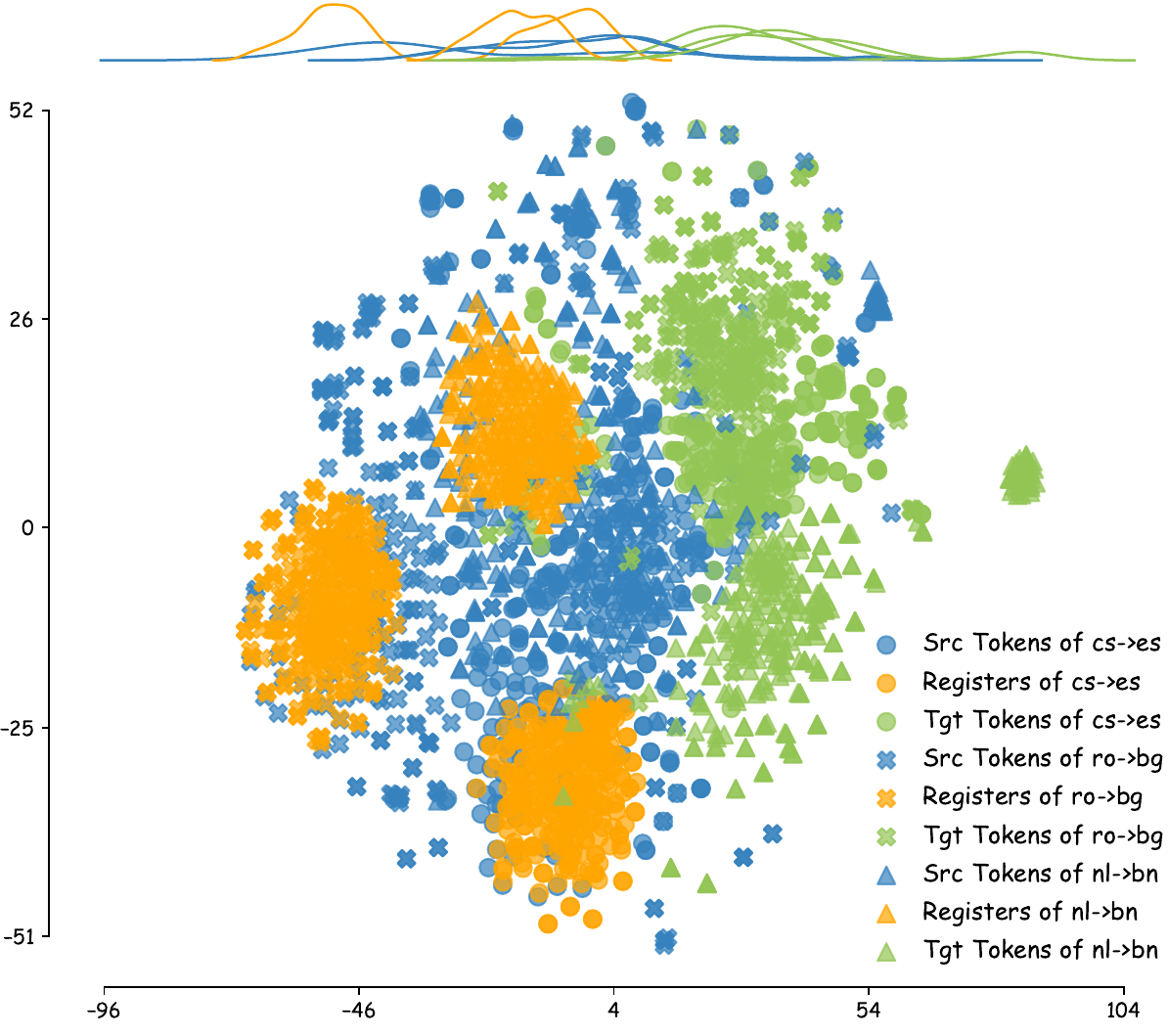}
        \caption{6th Layer}
        \label{fig:more_representation_3}
      \end{subfigure}
      \begin{subfigure}[b]{0.48\linewidth}
        \centering
        \includegraphics[width=\linewidth]{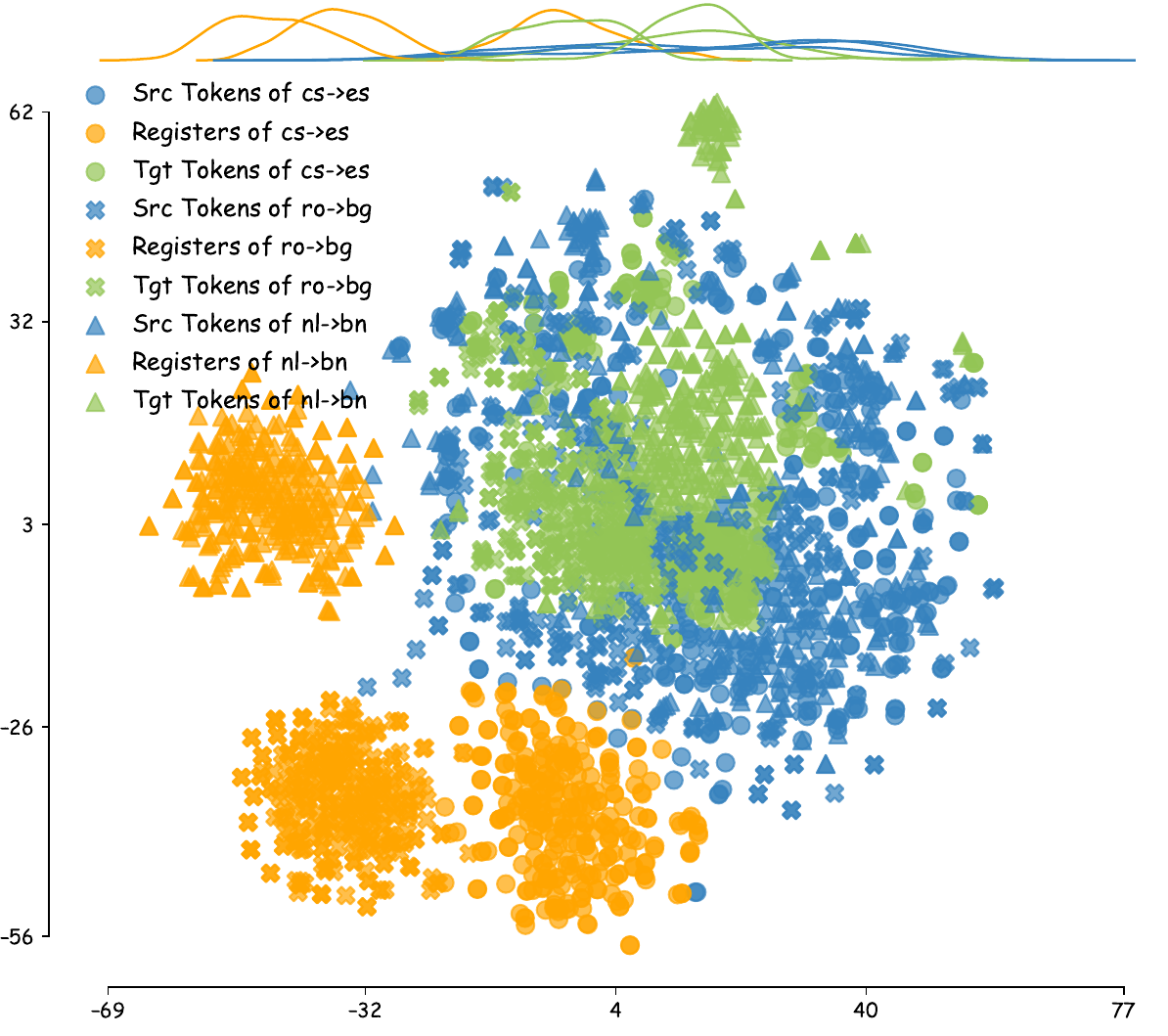}
        \caption{12th Layer}
        \label{fig:more_representation_4}
      \end{subfigure}
      \begin{subfigure}[b]{0.48\linewidth}
        \centering
        \includegraphics[width=\linewidth]{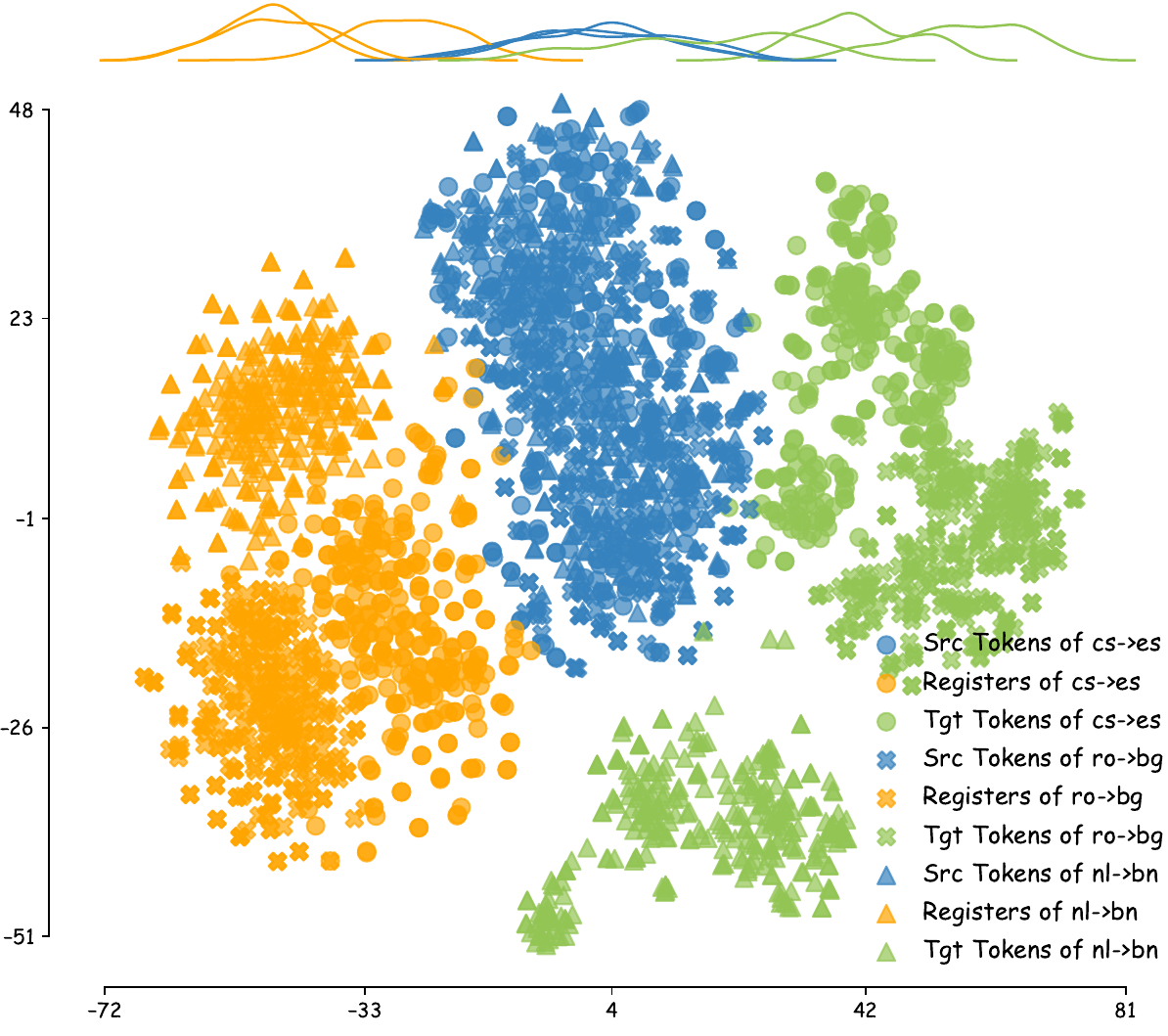}
        \caption{18th Layer}
        \label{fig:more_representation_5}
      \end{subfigure}
      \begin{subfigure}[b]{0.48\linewidth}
        \centering
        \includegraphics[width=\linewidth]{figures/scatter_24.pdf}
        \caption{24th Layer}
        \label{fig:more_representation_6}
      \end{subfigure}
    \caption{
    2D distributions of token-level representations extracted from the different layers of a model trained on EC-40.
    This illustration complements Figure \ref{fig:representation}.
    }
    \label{fig:more_representations}
\end{figure*}

We present Figure \ref{fig:more_representations} to supplement the analysis in Section \ref{section:analysis} on representation distributions, where Figure \ref{fig:representation} focuses specifically on the representation state in the 24th layer. Our observations are as follows:
(1) In the embedding layer and the 1st layer output (Figures \ref{fig:more_representation_1} and \ref{fig:more_representation_2}), source and target token representations are loosely distributed, while registers form three compact clusters based on language.
This is because registers lack semantic content and are distinguished only by positional encoding.  
(2) Starting from the 6th layer (Figure \ref{fig:more_representation_3}), source tokens begin to become distinguishable by language, and registers start to shift within the representation space toward the source tokens.
By the 12th layer (Figure \ref{fig:more_representation_4}), registers and source tokens are entirely separated in the representation space.  
(3) By the 18th layer (Figure \ref{fig:more_representation_5}), target tokens become clearly separated in the representation space, registers’ distribution becomes more diffuse, and the distribution of source tokens becomes more concentrated. These trends culminate in the state observed in the 24th layer (Figure \ref{fig:more_representation_6}), as described in Section \ref{section:analysis}.
These findings suggest two key phenomena: (1) registers progressively reinforce the semantic information they carry as they propagate through the layers; and (2) the representations of target tokens reflect their predicted state only in the higher layers.

\begin{table*}[t]
    \centering
    \resizebox{0.8\textwidth}{!}{
    \begin{tabular}{cclccccccccccc}
    \toprule
        ~  & ~ & ~ &  \multicolumn{2}{c}{High} & \multicolumn{2}{c}{Med} & \multicolumn{2}{c}{Low} & \multicolumn{2}{c}{Extra Low} & ~ & ~ & ~ \\ 
    \midrule
        ~ & \#params & Method & $\rightarrow$ & $\leftarrow$ & $\rightarrow$ & $\leftarrow$ & $\rightarrow$ & $\leftarrow$ & $\rightarrow$ & $\leftarrow$ & sup. & zero. & avg. \\ 
    \midrule
        \multirow{4}{*}{Enc-dec} & \multirow{3}{*}{242M} & vanilla & 23.27  & 25.61  & 20.43  & 24.61  & 16.39  & 14.25  & 17.73  & 13.34  & 49.27  & 18.16  & 19.68  \\ 
        ~ & ~ & \hspace{0.4em}+CL & 31.64  & 30.87  & 28.65  & 31.27  & 21.38  & 21.66  & 21.46  & 19.33  & 49.20  & 24.86  & 26.04  \\ 
        ~ & ~ & \hspace{0.4em}+LCS & 25.34  & 31.05  & 24.13  & 30.90  & 24.81  & 19.84  & 24.77  & 17.45  & \textbf{49.34}  & 23.70  & 24.95  \\
        ~ & 259M & \hspace{0.4em}+LAVS & 24.99 & 26.66 & 22.33 & 25.77 & 17.55 & 15.54 & 17.77 & 14.64 & 49.29 & 19.38 & 20.88 \\
    \cdashline{1-14}
        \multirow{3}{*}{Dec-only} & \multirow{3}{*}{217M} & vanilla & 27.26  & 27.10  & 24.42  & 27.16  & 18.30  & 18.01  & 18.46  & 16.17  & 48.44  & 21.00  & 22.34  \\ 
        ~ & ~ & \hspace{0.4em}+TDO & 30.25  & 29.87  & 25.87  & 29.61  & 20.61  & 19.75  & 20.50  & 18.00  & 48.92  & 23.32  & 24.57  \\ 
        ~ & ~ & \hspace{0.4em}+Ours & \textbf{33.74}  & \textbf{33.55}  & \textbf{31.77}  & \textbf{32.47}  & \textbf{24.87}  & \textbf{24.50}  & \textbf{24.82}  & \textbf{24.68}  & 48.87  & \textbf{28.00}  & \textbf{29.02}  \\ 
    \cdashline{1-14}
        \multirow{4}{*}{Enc-dec} & \multirow{3}{*}{418M} & vanilla & 27.91  & 31.84  & 25.40  & 31.48  & 20.57  & 16.02  & 20.79  & 15.33  & 50.11  & 22.60  & 23.95  \\ 
        ~ & ~ & \hspace{0.4em}+CL & 33.02  & 32.52  & 29.85  & 32.84  & 20.99  & 22.05  & 21.84  & 18.30  & \textbf{50.41}  & 25.51  & 26.73  \\ 
        ~ & ~ & \hspace{0.4em}+LCS & 24.67  & 33.81  & 23.88  & 32.99  & 26.30  & 18.10  & 25.71  & 16.87  & 50.16  & 24.26  & 25.57  \\
        ~ & 430M & \hspace{0.4em}+LAVS & 29.69 & 34.49 & 27.68 & 33.90 & 22.66 & 18.62 & 23.75 & 15.77 & 49.89 & 25.02 & 26.18\\
    \cdashline{1-14}
        \multirow{3}{*}{Dec-only} & \multirow{3}{*}{368M} & vanilla & 31.01  & 31.96  & 28.04  & 32.20  & 22.05  & 20.13  & 22.02  & 18.84  & 49.76  & 24.82  & 26.04  \\ 
        ~ & ~ & \hspace{0.4em}+TDO & 32.12  & 32.22  & 28.43  & 32.24  & 21.84  & 20.85  & 22.25  & 19.32  & 50.04  & 25.23  & 26.44  \\ 
        ~ & ~ & \hspace{0.4em}+Ours & \textbf{35.24}  & \textbf{35.00}  & \textbf{33.31}  & \textbf{34.56}  & \textbf{26.41}  & \textbf{26.01}  & \textbf{26.22}  & \textbf{25.61}  & 49.69  & \textbf{29.53}  & \textbf{30.51}  \\ 
    \bottomrule
    \end{tabular}
    }
    \caption{
    Averaged chrF++ scores of results on EC-40.
    All notations and abbreviations follow Table \ref{tab:result_ec40}.
    }
    \label{tab:result_ec40_chrf}
\end{table*}

\begin{table*}[t]
    \centering
    \resizebox{0.8\textwidth}{!}{
    \begin{tabular}{cclccccccccccc}
    \toprule
        ~  & ~ & ~ &  \multicolumn{2}{c}{High} & \multicolumn{2}{c}{Med} & \multicolumn{2}{c}{Low} & \multicolumn{2}{c}{Extra Low} & ~ & ~ & ~ \\ 
    \midrule
        ~ & \#params & Method & $\rightarrow$ & $\leftarrow$ & $\rightarrow$ & $\leftarrow$ & $\rightarrow$ & $\leftarrow$ & $\rightarrow$ & $\leftarrow$ & sup. & zero. & avg. \\ 
    \midrule
        \multirow{4}{*}{Enc-dec} & \multirow{3}{*}{242M} & vanilla & 50.21  & 49.61  & 42.07  & 44.93  & 28.62  & 28.15  & 32.94  & 31.17  & 77.26  & 52.29  & 53.72  \\ 
        ~ & ~ & \hspace{0.4em}+CL & 55.33  & 52.42  & 46.33  & 47.60  & 30.00  & 31.42  & 34.17  & 34.40  & 77.33  & 56.75  & 57.93  \\ 
        ~ & ~ & \hspace{0.4em}+LCS & 50.27  & 52.22  & 43.19  & 47.73  & 31.88  & 30.19  & 37.09  & 32.50  & \textbf{77.52}  & 55.44  & 56.71  \\ 
         ~ & 259M & \hspace{0.4em}+LAVS & 52.89 & 51.57 & 44.30 & 46.13 & 28.81 & 28.71 & 32.97 & 32.16 & 77.46 & 54.64 & 56.08 
 \\
    \cdashline{1-14}
        \multirow{3}{*}{Dec-only} & \multirow{3}{*}{217M} & vanilla & 52.66  & 50.88  & 43.80  & 45.88  & 29.19  & 30.13  & 33.88  & 32.64  & 77.10  & 54.41  & 55.71  \\ 
        ~ & ~ & \hspace{0.4em}+TDO & 54.39  & 52.48  & 45.22  & 47.40  & 30.09  & 30.76  & 34.53  & 33.59  & 77.13  & 56.16  & 57.36  \\ 
        ~ & ~ & \hspace{0.4em}+Ours & \textbf{56.94}  & \textbf{54.87}  & \textbf{48.74}  & \textbf{49.59}  & \textbf{32.61}  & \textbf{33.83}  & \textbf{37.10}  & \textbf{37.11}  & 77.12  & \textbf{60.23}  & \textbf{61.19}  \\ 
    \cdashline{1-14}
        \multirow{4}{*}{Enc-dec} & \multirow{3}{*}{418M} & vanilla & 54.06  & 54.76  & 45.71  & 49.32  & 31.11  & 29.68  & 35.37  & 32.49  & 78.48  & 56.84  & 58.08  \\ 
        ~ & ~ & \hspace{0.4em}+CL & 58.06  & 55.20  & 48.40  & 49.96  & 30.79  & 32.53  & 35.50  & 35.06  & \textbf{78.90}  & 59.26  & 60.38  \\ 
        ~ & ~ & \hspace{0.4em}+LCS & 51.52  & 55.63  & 44.84  & 50.74  & 33.84  & 30.80  & \textbf{38.91}  & 33.94  & 78.62  & 57.61  & 58.81  \\
        ~ & 430M & \hspace{0.4em}+LAVS & 54.18 & 56.30 & 46.24 & 50.73 & 33.02 & 31.60 & 37.64 & 32.46 & 76.29 & 57.90 & 58.95 \\
    \cdashline{1-14}
        \multirow{3}{*}{Dec-only} & \multirow{3}{*}{368M} & vanilla & 56.46  & 55.29  & 47.26  & 50.07  & 31.66  & 31.85  & 36.27  & 34.44  & 78.37  & 58.84  & 59.95  \\ 
        ~ & ~ & \hspace{0.4em}+TDO & 57.49  & 55.44  & 48.34  & 50.48  & 31.77  & 32.57  & 36.61  & 35.72  & 78.48  & 59.77  & 60.84  \\ 
        ~ & ~ & \hspace{0.4em}+Ours & \textbf{58.98}  & \textbf{56.92}  & \textbf{50.66}  & \textbf{51.82}  & \textbf{33.91}  & \textbf{35.37}  & 38.60  & \textbf{38.04}  & 78.12  & \textbf{62.66}  & \textbf{63.54}  \\
    \bottomrule
    \end{tabular}
    }
    \caption{
    Averaged COMET scores of results on EC-40.
    All notations and abbreviations follow Table \ref{tab:result_ec40}.
    }
    \label{tab:result_ec40_comet}
\end{table*}

\begin{table*}[t]
    \centering
    \resizebox{0.8\textwidth}{!}{
    \begin{tabular}{clcccccccccccccc}
    \toprule
        ~ & ~ & \multicolumn{2}{c}{English} & \multicolumn{2}{c}{Germanic} & \multicolumn{2}{c}{Romance} & \multicolumn{2}{c}{Slavic} & \multicolumn{2}{c}{Mal.-Polyn.} & \multicolumn{2}{c}{Asian} & ~ \\ 
    \midrule
        \multicolumn{2}{c}{Model} & $\rightarrow$ & $\leftarrow$ & $\rightarrow$ & $\leftarrow$ & $\rightarrow$ & $\leftarrow$ & $\rightarrow$ & $\leftarrow$ & $\rightarrow$ & $\leftarrow$ & $\rightarrow$ & $\leftarrow$ & avg. \\ 
    \midrule
        \multirow{3}{*}{M2M} & 483M & 50.43  & 54.36  & 44.84  & 46.24  & 43.96  & 48.26  & 43.36  & 43.06  & 37.54  & 41.77  & 39.83  & 27.85  & 42.53  \\ 
        ~ & 615M & 49.97  & 54.74  & 46.77  & 48.62  & 45.34  & 49.83  & 44.70  & 44.42  & 40.17  & 43.59  & 41.04  & 28.84  & 44.12  \\ 
        ~ & 1.2B & 54.80  & 54.44  & 49.02  & 47.06  & 47.49  & 48.04  & 45.87  & 43.35  & 32.78  & 40.68  & 30.90  & 28.01  & 42.56  \\ 
    \midrule
        \multirow{3}{*}{NLLB} & 615M & 55.54  & 61.73  & 48.48  & 50.39  & 47.38  & 51.54  & 46.38  & 45.34  & 45.98  & 50.96  & 42.56  & 29.73  & 46.70  \\ 
        ~ & 1.3B & 56.82  & 63.35  & 50.07  & 52.21  & 48.80  & 53.19  & 47.98  & 47.46  & 47.92  & 52.49  & 44.70  & 30.98  & 48.40  \\ 
        ~ & 3.3B & 57.88  & \textbf{64.27}  & 50.95  & 53.16  & 49.63  & 53.92  & 48.91  & 48.76  & 49.06  & \textbf{53.28}  & 45.71  & 31.97  & 49.35  \\
    \midrule
        \multirow{2}{*}{GPT} & 3.5 turbo & 55.30  & 61.60  & 49.39  & 52.16  & 48.31  & 53.34  & 47.54  & 46.35  & 45.64  & 48.05  & 43.46  & 31.20  & 47.41  \\ 
        ~ & 4o mini & 58.03  & 62.85  & 51.26  & 52.90  & 49.33  & 54.33  & 49.12  & 48.62  & \textbf{49.39}  & 52.25  & 45.86  & \textbf{34.11}  & 49.48  \\
    \midrule
        \multirow{2}{*}{MITRE} & 466M & \best58.11  & \best62.77  & \best51.40  & \best53.41  & \best50.06  & \best54.46  & \best49.18  & 48.62  & 47.83  & 52.04  & 45.10  & \better32.41  & 49.29  \\ 
        ~ & 913M & \best\textbf{58.84}  & \best64.01  & \best\textbf{52.40}  & \best\textbf{54.59}  & \best\textbf{51.03}  & \best\textbf{55.36}  & \best\textbf{50.32}  & \best\textbf{49.84}  & 48.97  & \best52.88  & \best\textbf{46.65}  & \better33.88  & \best\textbf{50.42}  \\ 
    \bottomrule
    \end{tabular}
    }
    \caption{
    Averaged chrF++ scores of results for comparing MITRE and baselines.
    All notations and abbreviations follow Table \ref{tab:result_large}.
    }
    \label{tab:result_large_chrf}
\end{table*}

\begin{table*}[t]
    \centering
    \resizebox{0.8\textwidth}{!}{
    \begin{tabular}{clcccccccccccccc}
    \toprule
        ~ & ~ & \multicolumn{2}{c}{English} & \multicolumn{2}{c}{Germanic} & \multicolumn{2}{c}{Romance} & \multicolumn{2}{c}{Slavic} & \multicolumn{2}{c}{Mal.-Polyn.} & \multicolumn{2}{c}{Asian} & ~ \\ 
    \midrule
        \multicolumn{2}{c}{Model} & $\rightarrow$ & $\leftarrow$ & $\rightarrow$ & $\leftarrow$ & $\rightarrow$ & $\leftarrow$ & $\rightarrow$ & $\leftarrow$ & $\rightarrow$ & $\leftarrow$ & $\rightarrow$ & $\leftarrow$ & avg. \\ 
    \midrule
        \multirow{3}{*}{M2M} & 483M & 81.63  & 81.40  & 78.90  & 77.03  & 80.48  & 78.49  & 79.85  & 80.31  & 68.34  & 72.75  & 78.67  & 78.57  & 77.74  \\ 
        ~ & 615M & 81.16  & 82.15  & 81.42  & 79.98  & 82.00  & 80.50  & 81.29  & 82.43  & 72.56  & 74.62  & 80.02  & 79.96  & 79.79  \\ 
        ~ & 1.2B & 85.93  & 85.17  & 84.15  & 82.87  & 84.46  & 83.12  & 83.87  & 85.41  & 75.91  & 77.83  & 82.95  & 82.58  & 82.66  \\ 
    \midrule
        \multirow{3}{*}{NLLB} & 615M & 86.61  & 86.76  & 84.06  & 82.77  & 84.50  & 83.05  & 83.41  & 84.39  & 81.26  & 83.51  & 82.12  & 82.02  & 83.33  \\ 
        ~ & 1.3B & 87.76  & 87.63  & 85.72  & 84.76  & 85.93  & 84.76  & 85.07  & 86.82  & 83.57  & 84.93  & 84.36  & 83.52  & 85.13  \\ 
        ~ & 3.3B & 88.22  & 88.09  & 86.49  & 85.58  & 86.58  & 85.45  & 85.84  & 87.96  & 84.63  & 85.45  & 85.42  & 84.56  & 85.96  \\ 
    \midrule
        \multirow{2}{*}{GPT} & 3.5 turbo & 87.67  & 88.02  & 86.26  & 85.80  & 86.62  & 85.96  & 85.91  & 87.50  & 83.09  & 82.09  & 85.45  & 85.77  & 85.66  \\ 
        ~ & 4o mini & \textbf{89.59} & \textbf{88.64}  & \textbf{87.50}  & \textbf{86.38}  & \textbf{87.58}  & \textbf{86.57}  & \textbf{87.08}  & \textbf{88.90}  & \textbf{85.89}  & \textbf{86.03}  & \textbf{86.99}  & \textbf{87.47}  & \textbf{87.16}  \\ 
    \midrule
        \multirow{2}{*}{MITRE} & 466M & 87.87  & 87.29  & 85.99  & 84.98  & 86.49  & 85.14  & 85.58  & 87.19  & 82.24  & 83.41  & 84.38  & 84.29  & 85.19  \\ 
        ~ & 913M & 88.11  & 87.81  & \better86.54  & \better85.61  & \better86.96  & \better85.70  & \better86.16  & \better88.03  & 83.15  & 83.80  & \better85.52  & \better85.35  & 85.88  \\ 
    \bottomrule
    \end{tabular}
    }
    \caption{
    Averaged COMET scores of results for comparing MITRE and baselines.
    All notations and abbreviations follow Table \ref{tab:result_large}.
    Given the different trend compared to Tables \ref{tab:result_large} and \ref{tab:result_large_chrf}, we not only mention it in Section \ref{section:pretrain_result}, but also provide an additional discussion in Appendix \ref{appendix:results_large}.
    }
    \label{tab:result_large_comet}
\end{table*}

\begin{table*}[t]
\renewcommand{\arraystretch}{1.85}
  \resizebox{\textwidth}{!}{
    \begin{tabular}{c|ccc|ccc|ccc|ccc|ccc}
    \toprule
      & \multicolumn{3}{c|}{Germanic} & \multicolumn{3}{c|}{Romance} & \multicolumn{3}{c|}{Slavic} & \multicolumn{3}{c|}{Indo-Aryan} & \multicolumn{3}{c}{Afro-Asiatic} \\
      & code & Language & Script & code & Language & Script & code & Language & Script & code & Language & Script & code & Language & Script \\
    \midrule
      \multirow{2}{*}{\begin{tabular}[c]{@{}c@{}}High\\ (5 million)\end{tabular}} & de & German & Latin & fr & French & Latin & ru & Russian & Cyrillic & hi & Hindi & Devanagari & ar & Arabic & Arabic \\
      & nl & Dutch & Latin & es & Spanish & Latin & cs & Czech & Latin & bn & Bengali & Bengali & he & Hebrew & Hebrew \\
    \midrule
      \multirow{2}{*}{\begin{tabular}[c]{@{}c@{}}Med\\ (1 million)\end{tabular}} & sv & Swedish & Latin & it & Italian & Latin & pl & Polish & Latin & kn & Kannada & Devanagari & mt & Maltese & Latin \\
      & da & Danish & Latin & pt & Portuguese & Latin & bg & Bulgarian & Cyrillic & mr & Marathi & Devanagari & ha & Hausa$^{*}$ & Latin \\
    \midrule
      \multirow{2}{*}{\begin{tabular}[c]{@{}c@{}}Low\\ (100 thousand)\end{tabular}} & af & Afrikaans & Latin & ro & Romanian & Latin & uk & Ukrainian & Cyrillic & sd & Sindhi & Arabic & ti & Tigrinya & Ethiopic \\
      & lb & Luxembourgish & Latin & oc & Occitan & Latin & sr & Serbian & Latin & gu & Gujarati & Devanagari & am & Amharic & Ethiopic \\
    \midrule
      \multirow{2}{*}{\begin{tabular}[c]{@{}c@{}}Extra Low\\ (50 thousand)\end{tabular}} & no & Norwegian & Latin & ast & Asturian & Latin & be & Belarusian & Cyrillic & ne & Nepali & Devanagari & kab & Kabyle$^{*}$ & Latin \\
      & ic & Icelandic & Latin & ca & Catalan & Latin & bs & Bosnian & Latin & ur & Urdu & Arabic & so & Somali & Latin \\
    \bottomrule
    \end{tabular}
  }
  \caption{Details of non-English languages in EC-40. This table is duplicated from \citet{zero-2023}. Numbers in the table represent the number of sentences paired to the English. Two exceptions are Hausa and Kabyle, where their data sizes are 334,000 and 18,448, respectively. }
  \label{tab:ec40}
\end{table*}

\end{document}